\documentclass[11pt,onecolumn]{article}
\usepackage{amsmath,amsthm,amssymb,amsfonts,latexsym,epsfig,graphicx}

\usepackage{algorithm}
\usepackage{algorithmic}

\usepackage{color}
\usepackage{dsfont}
\usepackage{wrapfig}
\usepackage{stmaryrd}
\usepackage{url}

\usepackage[left=0.9in,top=0.9in,right=0.9in,bottom=1.0in,nohead,textheight=10in,footskip=0.3in]{geometry}

\newtheorem{theorem}{Theorem}
\newtheorem{corollary}[theorem]{Corollary}

\newtheorem{lemma}[theorem]{Lemma}

\def\myparagraph#1{\vspace{4pt}\noindent{\bf #1~~}}

\long\def\ignore#1{}
\def\calA{{\cal A}}
\def\calB{{\cal B}}
\def\calC{{\cal C}}
\def\calD{{\cal D}}
\def\calE{{\cal E}}
\def\calF{{\cal F}}
\def\calG{{\cal G}}
\def\calH{{\cal H}}
\def\calI{{\cal I}}
\def\calL{{\cal L}}
\def\calN{{\cal N}}
\def\calP{{\cal P}}

\def\calR{{\cal R}}
\def\calS{{\cal S}}
\def\calV{{\cal V}}
\def\calT{{\cal T}}
\def\calU{{\cal U}}
\def\calO{{\cal O}}
\def\calX{{\cal X}}
\def\calY{{\cal Y}}

\def\Pproper{{P'}}
\def\Pall{{P}}
\def\PallLeft{{S}}
\def\PallRight{{P}}

\newif\ifTR\TRtrue

\newcommand\blfootnote[1]{%
  \begingroup
  \renewcommand\thefootnote{}\footnote{#1}%
  \addtocounter{footnote}{-1}%
  \endgroup
}

\begin{document}

\def\myparagraph#1{\vspace{2pt}\noindent{\bf #1~~}}



\long\def\ignore#1{}
\def\myps[#1]#2{\includegraphics[#1]{#2}}
\def\etal{{\em et al.}}
\def\Bar#1{{\bar #1}}
\def\br(#1,#2){{\langle #1,#2 \rangle}}
\def\setZ[#1,#2]{{[ #1 .. #2 ]}}
\def\Pr{\mbox{\rm Pr}}
\def\zd{,\ldots,}

\newcommand{\Z}{\mbox{$\mathbb Z$}}
\newcommand{\R}{\mbox{$\mathbb R$}}
\newcommand{\Q}{\mbox{$\mathbb Q$}}
\newcommand{\Qnn}{\mbox{$\mathbb Q_{\geq 0}$}}
\newcommand{\Qnnc}{\mbox{$\overline{\mathbb Q}_{\geq 0}$}}
\newcommand{\Qc}{\mbox{$\overline{\mathbb Q}$}}
\newcommand{\Rnn}{\mbox{$\mathbb R_{\geq 0}$}}
\newcommand{\Rnnc}{\mbox{$\overline{\mathbb R}_{\geq 0}$}}
\newcommand{\Rc}{\mbox{$\overline{\mathbb R}$}}
\newcommand{\N}{\mbox{$\mathbb N$}}

\def\closure#1{{\langle#1\rangle}}
\def\setof#1{{\left\{#1\right\}}}
\def\suchthat#1#2{\setof{\,#1\mid#2\,}} 
\def\event#1{\setof{#1}}
\def\q={\quad=\quad}
\def\qq={\qquad=\qquad}
\def\calA{{\cal A}}
\def\calC{{\cal C}}
\def\calD{{\cal D}}
\def\calE{{\cal E}}
\def\calF{{\cal F}}
\def\calG{{\cal G}}
\def\calI{{\cal I}}
\def\calH{{\cal H}}
\def\calL{{\cal L}}
\def\calN{{\cal N}}
\def\calP{{\cal P}}
\def\calR{{\cal R}}
\def\calS{{\cal S}}
\def\calT{{\cal T}}
\def\calU{{\cal U}}
\def\calV{{\cal V}}
\def\calO{{\cal O}}
\def\calX{{\cal X}}
\def\s{\footnotesize}
\def\calNG{{\cal N_G}}
\def\psfile[#1]#2{}
\def\psfilehere[#1]#2{}
\def\epsfw#1#2{\includegraphics[width=#1\hsize]{#2}}
\def\assign(#1,#2){\langle#1,#2\rangle}
\def\edge(#1,#2){(#1,#2)}
\def\VS{\calV^s}
\def\VT{\calV^t}
\def\slack(#1){\texttt{slack}({#1})}
\def\barslack(#1){\overline{\texttt{slack}}({#1})}
\def\NULL{\texttt{NULL}}
\def\PARENT{\texttt{PARENT}}
\def\GRANDPARENT{\texttt{GRANDPARENT}}
\def\TAIL{\texttt{TAIL}}
\def\HEADORIG{\texttt{HEAD$\_\:$ORIG}}
\def\TAILORIG{\texttt{TAIL$\_\:$ORIG}}
\def\HEAD{\texttt{HEAD}}
\def\CURRENTEDGE{\texttt{CURRENT$\!\_\:$EDGE}}

\def\unitvec(#1){{{\bf u}_{#1}}}
\def\uvec{{\bf u}}
\def\vvec{{\bf v}}
\def\Nvec{{\bf N}}

\newcommand{\bg}{\mbox{$\bf g$}}
\newcommand{\bh}{\mbox{$\bf h$}}

\newcommand{\bx}{\mbox{$x$}}
\newcommand{\by}{\mbox{\boldmath $y$}}
\newcommand{\bz}{\mbox{\boldmath $z$}}
\newcommand{\bu}{\mbox{\boldmath $u$}}
\newcommand{\bv}{\mbox{\boldmath $v$}}
\newcommand{\bw}{\mbox{\boldmath $w$}}
\newcommand{\bvarphi}{\mbox{\boldmath $\varphi$}}

\newcommand\myqed{{}}



\title{\Large\bf  \vspace{0pt} Inference algorithms for pattern-based CRFs on sequence data}

\author{
      Rustem Takhanov \hspace{125pt}  Vladimir Kolmogorov \hspace{5pt} ~\\ 
{\normalsize\tt takhanov@mail.ru} \hspace{150pt}   {\normalsize\tt vnk@ist.ac.at} \hspace{20pt} ~\\
\normalsize Nazarbayev University, Kazakhstan \hspace{40pt}
\normalsize Institute of Science and Technology Austria 
}

\begin{figure}
\vspace{0pt}
\center
\begin{eqnarray*}
\mbox{Accepted to {\em Algorithmica}. ~~The final publication is available at link.springer.com. }\\
\mbox{\url{http://link.springer.com/article/10.1007/s00453-015-0017-7}.~~~~~~~~ }
\end{eqnarray*}
\vspace{-10pt}
\end{figure}

\date{}
\maketitle

\blfootnote{A preliminary version of this paper appeared in 
Proceedings of the 30th International Conference on Machine Learning (ICML), 2013~\cite{TK:ICML13}. This work was partially supported by the European Research Council under the European Unions Seventh Framework Programme (FP7/2007-2013)/ERC grant agreement no 616160.}

\begin{abstract}
We consider {\em Conditional Random Fields (CRFs) with pattern-based potentials}
defined on a chain. In this model the energy of a string (labeling) $x_1\ldots x_n$
is the sum of terms over intervals $[i,j]$ where each term is non-zero only if the substring $x_i\ldots x_j$
equals a prespecified pattern $\alpha$. Such CRFs can be naturally applied to many sequence tagging problems.

We present efficient algorithms for the three standard inference tasks in a CRF, namely
computing (i) the partition function, (ii) marginals, and (iii) computing the MAP.
Their complexities are respectively
$O(n L)$, $O(n L \ell_{\max})$ and
$O(n L \min\{|D|,\log (\ell_{\max}\!+\!1)\})$
where $L$ is the combined length of input patterns, $\ell_{\max}$ is the maximum length of a pattern,
and $D$ is the input alphabet.
This improves on the previous algorithms of [Ye et al. NIPS 2009] whose complexities are respectively $O(n L |D|)$,
$O\left(n |\Gamma| L^2 \ell_{\max}^2\right)$ and $O(n L |D|)$,
where $|\Gamma|$ is the number of input patterns. In addition, we give an efficient algorithm for sampling,
 and revisit the case of MAP with non-positive weights.
\end{abstract}

This paper addresses the {\em sequence labeling} (or the {\em sequence tagging}) problem: given an observation
$z$ (which is usually a sequence of $n$ values), infer labeling $x=x_1\ldots x_n$
where each variable $x_i$ takes values in some finite domain $D$.
Such problem appears in many domains such as text and speech analysis, signal analysis, and bioinformatics.

One of the most successful approaches for tackling the problem is the Hidden Markov Model (HMM).
The $k$th order HMM is given by the probability distribution $p(x|z)=\frac{1}{Z}\exp\{-E(x|z)\}$
with the energy function
\begin{equation}
E(x|z)=\sum_{i\in [1,n]} \psi_i(x_i,z_i)+\sum_{(i,j)\in \calE_k}\psi_{ij}(x_{i:j})
\label{eq:HMM}
\end{equation}
where  $\calE_k=\{(i,i+k)\:|\:i\in [1,n-k]\}$ and $x_{i:j}=x_i\ldots x_j$ is the substring of $x$ from $i$ to $j$.
A popular generalization is the Conditional Random Field model~\cite{Lafferty:ICML01} that allows all terms to depend
on the full observation $z$:
\begin{equation}
E(x|z)=\sum_{i\in [1,n]} \psi_i(x_i,z)+\sum_{(i,j)\in \calE_k}\psi_{ij}(x_{i:j},z)
\label{eq:CRF}
\end{equation}
We study a particular variant of this model called a {\em pattern-based CRF}.
It is defined via
\begin{equation}
E(x|z)=\sum_{\alpha\in\Gamma}\sum_{\substack{[i,j]\subseteq[1,n]\\j-i+1=|\alpha|}}\psi^\alpha_{ij}(z)\cdot[x_{i:j}=\alpha]
\label{eq:patternCRF}
\end{equation}
where $\Gamma$ is a fixed set of non-empty words, $|\alpha|$ is the length of word $\alpha$ and $[\cdot]$ is the {\em Iverson bracket}.
If we take $\Gamma=D\cup D^k$ then~\eqref{eq:patternCRF}
becomes equivalent to~\eqref{eq:CRF}; thus we do not loose generality (but gain more flexibility).

Intuitively, pattern-based CRFs allow to model long-range interactions for selected subsequences of labels.
This could be useful for a variety of applications:
in part-of-speech tagging patterns could correspond to certain syntactic constructions or stable idioms;
in protein secondary structure prediction - to sequences of dihedral angles corresponding to stable configuration such as $\alpha$-helixes;
in gene prediction - to sequences of nucleatydes with supposed functional roles such as ``exon'' or ``intron'', specific codons, etc.

\noindent{\bf Inference~} This paper focuses on inference algorithms for pattern-based CRFs.
The three standard inference tasks are
\begin{itemize}
\item computing the partition function $Z=\sum_x \exp\{-E(x|z)\}$; 
\item  computing marginal probabilities $p(x_{i:j}=\alpha|z)$ for all triplets $(i,j,\alpha)$ present in~\eqref{eq:patternCRF}; 
\item  computing MAP, i.e.\ minimizing energy~\eqref{eq:patternCRF}. 
\end{itemize}
The complexity of solving these tasks is discussed below.
We denote $L=\sum_{\alpha\in\Gamma}|\alpha|$ to be total length of patterns
and $\ell_{\max}=\max_{\alpha\in\Gamma}|\alpha|$ to be the maximum length of a pattern.

A naive approach is to use standard message passing techniques for an HMM of order $k\!=\!\ell_{\max}-1$.
However, they would take $O(n|D|^{k+1})$ time which would become impractical for large $k$.
More efficient algorithms with complexities $O(n L |D|)$,
$O\left(n |\Gamma|L^2 \ell_{\max}^2\right)$ and $O(n L |D|)$ respectively
were given by Ye et al. \cite{Ye:NIPS09}.\footnote{Some of the bounds stated in \cite{Ye:NIPS09} are actually weaker.
However, it is not difficult to show that their algorithms can be implemented in times stated above, using our Lemma~\ref{lemma:varphis}.}
Our first contribution is to improve this to
$O(n L)$, $O(n L \ell_{\max})$ and
$O(n L\cdot \min\{|D|,\log (\ell_{\max}+1)\})$
respectively (more accurate estimates are given in the next section).

We also give an algorithm for sampling from the distribution $p(x|z)$.
Its complexity is either (i) $O(nL)$ per sample, or (ii)
 $O(n)$ per sample with an $O(nL|D|)$ preprocessing
(assuming that we have an oracle that produces independent samples from the uniform distribution on $[0,1]$
in $O(1)$ time).

Finally, we consider the case when all costs $\psi^\alpha_{ij}(z)$ are non-positive.
Komodakis and Paragios \cite{Komodakis:CVPR09} gave an $O(n L)$ technique for minimizing energy~\eqref{eq:patternCRF} in this case.
We present a modification that has the same worst-case complexity but can beat
the algorithm in \cite{Komodakis:CVPR09} in the best case.

\noindent {\bf Related work} The works of~\cite{Ye:NIPS09} and \cite{Komodakis:CVPR09}
 are probably the most related to our paper.
The former applied pattern-based CRFs to the handwritten character recognition problem and to the problem of identification of named entities from texts.
The latter considered a pattern-based CRF on a grid for a computer vision application;
the MAP inference problem in~\cite{Komodakis:CVPR09} was converted to sequence labeling problems by decomposing the grid into thin ``stripes''.

Qian et al. \cite{Qian:ICML09} considered a more general formulation in which a single pattern is characterized
by a set of strings rather than a single string $\alpha$. They proposed an exact inference algorithm
and applied it to the OCR task and to the Chinese Organization Name Recognition task.
However, their algorithm could take time exponential in the total lengths of input patterns;
no subclasses of inputs were identified which could be solved in polynomial time.

A different generalization (for non-sequence data) was proposed by Rother et al. \cite{Rother:CVPR09}. Their inference procedure
reduces the problem to the MAP estimation in a  pairwise CRF with cycles,
which is then solved with approximate techniques such as BP, TRW or QPBO.
This model was applied to the texture restoration problem.

Nguyen et al. \cite{Nguyen:11} extended algorithms in~\cite{Ye:NIPS09} to the {\em Semi-Markov model} \cite{Sarawagi:NIPS04}.
We conjecture that our algorithms can be extended to this case as well, and can yield a better complexity compared to \cite{Nguyen:11}.

In~\cite{TK:ICML13} we applied the pattern-based CRF model to the problem of the protein dihedral angles prediction.

\section{Notation and preliminaries}
First, we introduce a few definitions.
\begin{itemize}
\item A {\em pattern} is a pair $\alpha=([i,j],x)$ where $[i,j]$ is an interval in $[1,n]$ and $x=x_i\ldots x_j$ is
a sequence over alphabet $D$ indexed by integers in $[i,j]$ ($j\ge i-1$).
The {\em length} of $\alpha$ is denoted as   $|\alpha|=|x|= j-i+1$.
\item Symbols ``$\ast$'' denotes an arbitrary word or pattern
(possibly the empty word $\varepsilon$ or the empty pattern $\varepsilon_s\triangleq([s+1,s],\varepsilon)$ at position $s$).
The exact meaning will always be clear from the context.
Similary, ``+'' denotes an arbitrary non-empty word or pattern.
\item The concatenation of patterns $\alpha=([i,j],x)$ and $\beta=([j+1,k],y)$ is the
pattern $\alpha\beta\triangleq([i,k],xy)$.
Whenever we write $\alpha\beta$ we assume that it is defined, i.e.\ $\alpha=([\cdot,j],\cdot)$ and $\beta=([j+1,\cdot],\cdot)$ for some $j$.
\item For a pattern $\alpha=([i,j],x)$ and interval $[k,\ell]\subseteq[i,j]$,
the {\em subpattern of $\alpha$ at position $[k,\ell]$} is the pattern $\alpha_{k:\ell}\triangleq([k,\ell],x_{k:\ell})$
where $x_{k:\ell}=x_k\ldots x_\ell$. \\
If $k=i$ then $\alpha_{k:\ell}$ is called a {\em prefix} of $\alpha$.
If $\ell=j$ then $\alpha_{k:\ell}$ is a {\em suffix} of $\alpha$.
\item If $\beta$ is a subpattern of $\alpha$, i.e.\ $\beta=\alpha_{k:\ell}$ for some $[k,\ell]$,
then we say that $\beta$ is {\em contained} in $\alpha$. 
This is equivalent to the condition $\alpha=\ast\beta\ast$.
\item $D^{i:j}=\{([i,j],x)\:|\:x\in D^{[i,j]}\}$ is the
set of patterns with interval $[i,j]$.
We typically use letter $x$ for patterns in $D^{1:s}$ and letters $\alpha,\beta,\ldots$ for other patterns.
Patterns $x\in D^{1:s}$ will be called {\em partial labelings}.
\item For a set of patterns $\Pi$ and index $s\in[0,n]$ we denote $\Pi_s$ to be the set of patterns in $\Pi$
that end at position $s$: $\Pi_s=\{([i,s],\alpha)\in\Pi\}$.
\item
For a pattern $\alpha$ let $\alpha^-$ be the
prefix of $\alpha$ of length $|\alpha|-1$; if $\alpha$ is empty
then $\alpha^-$ is undefined.
\end{itemize}
We will consider the following general problem.
Let $\Pi^\circ$ be the set of patterns of words in $\Gamma$ placed at all possible positions:
$\Pi^\circ=\{([i,j],\alpha)\:|\:\alpha\in\Gamma)\}$.
Let $(R,\oplus,\otimes)$ be a commutative semiring with elements $\mathbb{O},\mathds{1}\in R$
which are identities for $\oplus$ and $\otimes$ respectively. Define the cost of pattern $x\in D^{i:j}$ via
\begin{equation}
\!\!\!\!\!\!f( x ) =\!\!\!\!\! \bigotimes_{\alpha\in\Pi^\circ,x=\ast\alpha\ast} \!\!\! c_\alpha
\label{eq:F}
\end{equation}
where $c_\alpha\!\in\! R$ are fixed constants. (Throughout the paper we adopt the convention that operations $\oplus$ and $\otimes$ over the empty set of arguments give
respectively $\mathbb{O}$ and $\mathds{1}$, and so e.g. $f(\varepsilon_s)=\mathds{1}$.)
Our goal is to compute\!\!\!\!\!\!
\begin{equation}
Z=\bigoplus_{x\in D^{1:n}} f(x)\label{eq:Fsum}
\end{equation}

\noindent
{\bf Example 1~} {\em If $(R,\oplus,\otimes)=(\mathbb R,+,\times)$ then problem~\eqref{eq:Fsum} is equivalent
to computing the partition function 
for the energy~\eqref{eq:patternCRF},
if we set $c_{([i,j],\alpha)}=\exp\{- \psi^\alpha_{ij}(z) \}$. }

\noindent
{\bf Example 2~} {\em If $(R,\oplus,\otimes)=(\overline {\mathbb R},\min,+)$ where
$\overline {\mathbb R}\triangleq \mathbb R\cup\{+\infty\}$
then we get the problem of minimizing energy~\eqref{eq:patternCRF},
if  $c_{([i,j],\alpha)}=\psi^\alpha_{ij}(z)$. }

The complexity of our algorithms will be stated in terms of the following quantities:
\begin{itemize}
\item $\Pall=|\{\alpha\:|\:\exists\alpha\ast\in\Gamma,\alpha\ne\varepsilon\}|$ is the number of distinct non-empty prefixes of words in $\Gamma$.
Note that $\Pall\le L$.
\item $\Pproper=|\{\alpha\:|\:\exists\alpha+\in\Gamma\}|$ is the number of distinct proper prefixes of words in $\Gamma$.
There holds $\frac{\Pall}{\Pproper}\in[1,|D|]$. \\
If $\Gamma=D^1\cup D^2\cup\ldots\cup D^k$ then
$\frac{\Pall}{\Pproper}=|D|$. If
$\Gamma$ is a sparse random
subset of the set above then
 $\frac{\Pall}{\Pproper}\approx 1$.
\item $
I(\Gamma)=\{\alpha\:|\:\exists\alpha\ast,\ast\alpha\in\Gamma,\alpha\ne\varepsilon\}
$
is the set of non-empty words which are both prefixes and suffixes of some words in $\Gamma$.
Note that $\Gamma \!\subseteq\! I(\Gamma)$ and
 $|I(\Gamma)|\!\le\! \Pall$.
\end{itemize}
We will present 6 algorithms:

\noindent
{\bf Sec.~\ref{sec:ring}}: $\Theta(n\Pall)$ algorithm for the case
when $(R,\oplus,\otimes)$ is a ring, i.e.\ it has operation $\ominus$ that satisfies
$\left(a\ominus b\right)\oplus b = a$ for all $a,b\in R$.
This holds for the semiring in Example 1 (but not for Example 2).
\\
{\bf Sec.~\ref{sec:sampling}}: $\Theta(n\Pall)$ algorithm for sampling.
Alternatively, it can be implemented to produce independent samples in $\Theta(n)$ time per
sample with a $\Theta(n\Pall|D|)$ preprocessing.
\\
{\bf Sec.~\ref{sec:marginals}}: $O(n\sum_{\alpha\in I(\Gamma)}|\alpha|)$ algorithm for computing marginals for all patterns $\alpha\in\Pi^\circ$.
\\
{\bf Sec.~\ref{sec:semiring:D}}: $\Theta(n\Pproper|D|)$ algorithm
for a general commutative semiring, which is equivalent to the algorithm in~\cite{Ye:NIPS09}.
It will be used as a motivation for the next algorithm. \\
{\bf Sec.~\ref{sec:semiring:logL}}: $O(n\Pall\log \Pall)$ algorithm
for a general commutative semiring; for the semiring in Example 2 the complexity
can be reduced to $O(n\Pall\log (\ell_{\max}+1))$. \\
{\bf Sec.~\ref{sec:negative}}: $O(n\Pall)$ algorithm for the case
$(R,\oplus,\otimes)=(\overline{\mathbb R},\min,+)$, $c_\alpha\le 0$ for all $\alpha\in\Pi^\circ$.

All algorithms will have the following structure. Given the set of input patterns $\Pi^\circ$, we first
construct another set of patterns $\Pi$; it will
typically be either the set of prefixes or the set of proper prefixes of patterns in $\Pi^\circ$.
This can be done in a preprocessing step since
sets $\Pi_s$ will be isomorphmic (up to a shift) for indexes $s$ that
are sufficiently far from the boundary. (Recall that $\Pi_s$ is the set of patterns in $\Pi$ that end at position $s$.)
Then we recursively compute {\em messages} $M_s(\alpha)$ for $\alpha\in\Pi_s$
which have the following interpretation: $M_s(\alpha)$ is the sum (``$\oplus$'')
of costs $f(x)$ over a certain set of partial labelings of the form $x=\ast \alpha\in D^{1:s}$.
In some of the algorithms we also compute
messages $W_s(\alpha)$ which is the sum of $f(x)$ over {\bf all} partial labelings of the form $x=\ast \alpha\in D^{1:s}$.

\begin{figure}[t]
\vskip 0.0in
\small
\begin{center}
\begin{tabular}{c@{\hspace{2pt}}c@{\hspace{1pt}}c@{\hspace{0pt}}c@{\hspace{0pt}}c@{\hspace{0pt}}c}
.&.&.&.&.&. \\
.&.&.&.&.&0 \\
.&.&.&.&.&1 \\
.&.&.&.&1&0 \\
.&.&.&1&0&0 \\
.&.&.&1&0&1 \\
.&.&1&0&0&0 \\
.&.&1&0&1&0
\end{tabular}
\normalsize
\hspace{0pt}
\begin{tabular}{c}
\includegraphics[scale=0.20]{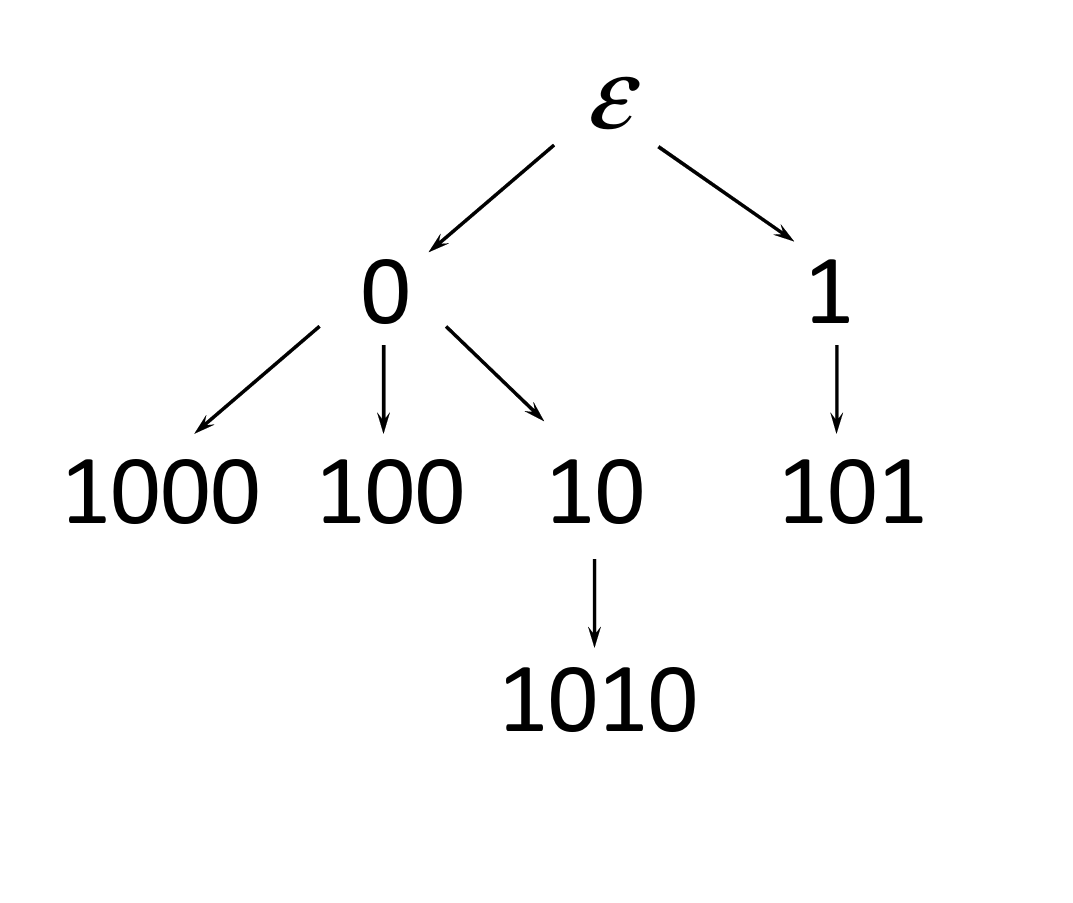} \vspace{-11pt} \\
\end{tabular}
\end{center}
\caption{Graph $G[\Pi_s]$ for the set of 8 patterns shown on the left (for brevity, their intervals are not shown;
they all end at the same position $s$.) This set of patterns would arise if $\Gamma=\{0,1,1000,1010\}$ and
$\Pi$ was defined as the set of all prefixes of patterns in $\Pi^\circ$.
}
\label{fig:graph}
\vskip -0.1in
\end{figure}

\myparagraph{Graph $G[\Pi_s]$} The following construction will be used throughout the paper. Given a set of patterns $\Pi$
and index $s$, we define $G[\Pi_s]=(\Pi_s,E[\Pi_s])$ to
be the Hasse diagram of the partial order $\preceq$ on $\Pi_s$, where
 $\alpha\preceq\beta$ iff $\alpha$ is a suffix of $\beta$ ($\beta=\ast\alpha$).
In other words, $G[\Pi_s]$ is a directed acyclic graph with the following set of edges:
 $(\alpha,\beta)$ belongs to $E[\Pi_s]$ for $\alpha,\beta\in \Pi_s$ if $\alpha\prec\beta$ and
there exists no ``intermediate'' pattern $\gamma\in\Pi_s$ with $\alpha\prec\gamma\prec\beta$.
It can be checked that graph $G[\Pi_s]$ is a directed forest.
If $\varepsilon_s\in \Pi_s$ then $G[\Pi_s]$ is connected and therefore is a tree. In this case we treat $\varepsilon_s$ as the root.
An example is shown in Fig.~\ref{fig:graph}.

\myparagraph{Computing partial costs} Recall that $f(\alpha)$ for a pattern $\alpha$
is the cost of all patterns inside $\alpha$ (eq.~\eqref{eq:F}).
We also define $\phi(\alpha)$ to be the cost of only those patterns that are suffixes of $\alpha$:
\begin{equation}
\phi( \alpha ) = \bigotimes_{\beta\in\Pi^\circ,\alpha=\ast\beta}  c_\beta
\end{equation}
Quantities
$\phi(\alpha)$ and $f(\alpha)$ will be heavily used by the algorithms below; let us show
how to compute them efficiently.
\begin{lemma}
Let $\Pi$ be a set of patterns with $\varepsilon_s\in \Pi$ for all $s\in[0,n]$.
Values $\phi(\alpha)$ for all $\alpha\in\Pi$
can be computed using $O(|\Pi|)$ multiplications (``$\otimes$'').
The same holds for values $f(\alpha)$ assuming that $\Pi$ is prefix-closed,
i.e.\ $\alpha^-\in\Pi$ for all non-empty patterns $\alpha\in\Pi$.
\label{lemma:varphis}
\end{lemma}
\ifTR
\begin{proof} To compute $\phi(\cdot)$ for patterns $\alpha\in\Pi_s$, we use the following procedure: (i) set $\phi(\varepsilon_s):=\mathds{1}$;
(ii) traverse edges $(\alpha,\beta)\in E[\Pi_s]$ of tree
$G[\Pi_s]$ (from the root to the leaves) and
set
$$
\phi(\beta):=\begin{cases}
\phi(\alpha)\otimes c_\beta & \mbox{if }\beta\in\Pi^\circ \\
\phi(\alpha)  & \mbox{otherwise}
\end{cases}
$$
Now suppose that $\Pi$ is prefix-closed.
After computing $\phi(\cdot)$, we go through indexes $s\in[0,n]$ and set
$$
f(\varepsilon_s)\!:=\!\mathds{1},\qquad f(\alpha)\!:=\!f(\alpha^-)\otimes \phi(\alpha)\quad\forall \alpha\in \Pi_s-\{\varepsilon_s\}
$$
\end{proof}
\else
\fi

\myparagraph{Sets of partial labelings}
Let $\Pi_s$ be a set of patterns that end at position $s$.
 Assume that $\varepsilon_s\in\Pi_s$. For a pattern $\alpha\in \Pi_s$
we define
\begin{eqnarray}
\calX_s(\alpha)&=&\{x\in D^{1:s}\:|\:x=\ast\alpha\} \\
\calX_s(\alpha;\Pi_s)&=&\calX_s(\alpha)-\bigcup_{(\alpha,\beta)\in E[\Pi_s]} \calX_s(\beta)
\end{eqnarray}
It can be seen that sets $\calX_s(\alpha;\Pi_s)$ are disjoint, and their union over $\alpha\in \Pi_s$ is $D^{1:s}$.
Furthermore, there holds
\begin{eqnarray}
\calX_s(\alpha;\Pi_s)=\{x\in \calX_s(\alpha)\:|\:x\ne\ast\beta\;\;\forall \beta=+\alpha\in \Pi_s\}
\label{eq:calXs'}
\end{eqnarray}
We will use eq.~\eqref{eq:calXs'} as the definition of $\calX_s(\alpha;\Pi_s)$ in the case when $\alpha\notin \Pi_s$.


\section{Computing partition function}\label{sec:ring}
In this section we give an algorithm for computing quantity~\eqref{eq:Fsum}
assuming that $(R,\oplus,\otimes)$ is a ring. This can be used, in particular, for computing the partition function.
We will assume that $D\subseteq\Gamma$; we can always add $D$ to $\Gamma$ if needed\footnote{Note
that we still claim complexity $O(n\Pall)$ where $\Pall$ is the number of distinct non-empty prefixes of words in the {\em original} set $\Gamma$.
Indeed, we can assume w.l.o.g.\ that each letter in $D$ occurs in at least one word $\alpha\!\in\!\Gamma$.
(If not, then we can ``merge'' non-occuring letters to a single
letter and add this letter to $\Gamma$; clearly, any instance over the original pair $(D,\Gamma)$ can be equivalenly formulated
as an instance over the new pair. The transformation increases $\Pall$ only by $1$). The assumption implies that $|D|\le \Pall$.
Adding $D$ to $\Gamma$ increases $\Pall$ by at most $\Pall$, and thus does not affect bound $O(n\Pall)$.}.

First, we select set $\Pi$ as
the set of prefixes of patterns in $\Pi^\circ$:
\begin{equation}
\Pi=\{\alpha\:|\:\exists\alpha\ast\in\Pi^\circ\}
\label{eq:GLADGAKGADF}
\end{equation}
We will compute the following quantities for each $s\in[0,n]$, $\alpha\in\Pi_s$: 
\begin{equation}
M_s(\alpha)=\!\!\!\!\!\!\bigoplus_{x\in\calX_s(\alpha;\Pi_s)}\!\!\!\!\!\! f(x)
\;,\quad
W_s(\alpha)=\!\!\!\!\bigoplus_{x\in\calX_s(\alpha)}\!\!\! f(x)
\label{eq:DP'':MandW}
\end{equation}

It is easy to see that for $\alpha\in \Pi_s$ the following equalities relate $M_s$ and $W_s$:
\begin{subequations}
\begin{eqnarray}
M_s(\alpha)&=&W_s(\alpha)\ominus\bigoplus_{(\alpha, \beta)\in E[\Pi_s]} W_s(\beta) \label{eq:alg'':M} \\
W_s(\alpha)&=&M_s(\alpha)\oplus\bigoplus_{(\alpha, \beta)\in E[\Pi_s]} W_s(\beta) \label{eq:alg'':W}
\end{eqnarray}
\end{subequations}
These relations motivate the following algorithm.
Since $|\Pi_s|=\Pall+1$ for indexes $s$ that are sufficiently far from the boundary, its complexity is $\Theta(n\Pall)$
assuming that values $\phi(\alpha)$ in eq.~\eqref{eq:alg'':Mupdate} are computed using Lemma~\ref{lemma:varphis}.

\begin{algorithm}
\caption{Computing $Z=\bigoplus\nolimits_{x\in D^{1:n}} f(x)$ for a ring}\label{alg:DP''}
\begin{algorithmic}[1]
\STATE initialize messages:
set 
$W_0(\varepsilon_0):=\mathbb{O}$
\STATE for $s=1,\ldots,n$ traverse nodes $\alpha\in \Pi_s$ of tree $G[\Pi_s]$
starting from the leaves and set
\begin{subequations}
\begin{eqnarray}
M_{s}( \alpha ) &:=& \phi(\alpha)\otimes\left[ W_{s-1}(\alpha^-) \ominus \bigoplus_{(\alpha,\beta)\in E[\Pi_s]} W_{s-1}(\beta^{-}) \right]
\label{eq:alg'':Mupdate} \\
W_{s}( \alpha ) &:=& M_{s}( \alpha ) \oplus \bigoplus_{(\alpha,\beta)\in E[\Pi_s]} W_s(\beta)
\label{eq:alg'':Wupdate}
\end{eqnarray}
\end{subequations}
Exception: if $\alpha=\varepsilon_s$ then set $M_s(\alpha):=\mathbb{O}$ instead of~\eqref{eq:alg'':Mupdate}
\STATE return
$
Z:=W_n(\varepsilon_n)
$
\end{algorithmic}
\end{algorithm}

\begin{theorem} Algorithm~\ref{alg:DP''} is correct, i.e.\ it
returns the correct value of $Z=\bigoplus_x F(x)$.
\label{th:correctnessDP''}
\end{theorem}

\ifTR
\subsection{Proof of Theorem~\ref{th:correctnessDP''}}
Eq.~\eqref{eq:alg'':Wupdate} coincides with \eqref{eq:alg'':W}; let us show
that eq.~\eqref{eq:alg'':Mupdate} holds for any $\alpha\in \Pi_s-\{\varepsilon_s\}$.
(Note, for $\alpha=\varepsilon_s$ step 2 is correct:
assumption $D\subseteq\Gamma$ implies that $D^{s:s}\subseteq\Pi_s$, and therefore
$\calX_s(\varepsilon_s;\Pi_s)=\varnothing$, $M_s(\varepsilon_s)=\mathbb{O}$).

For a partial labeling $x\in D^{1:s}$ define the ``reduced partial cost'' as
\begin{equation}
\!\!\!\!\!\!f^-( x ) =\!\!\!\!\! \bigotimes_{\alpha\in\Pi^\circ,x=\ast\alpha+} \!\!\! c_\alpha
\end{equation}
It is easy to see from~\eqref{eq:DP'':MandW} that for any $\alpha\in\Pi_s-\{\varepsilon_s\}$
\begin{equation}
W_{s-1}(\alpha^-)=\sum_{x\in\calX_s(\alpha)} f^-(x)
\label{eq:DP'':Wminus}
\end{equation}
Consider $\alpha\in\Pi_s-\{\varepsilon_s\}$. We will show that
for any $x\in\calX_s(\alpha)$ there holds
\begin{equation}
\llbracket x \in \calX_s(\alpha;\Pi_s)\rrbracket\otimes f(x) = \phi(\alpha)\otimes \left[ f^-(x) \ominus
\bigoplus_{\substack{(\alpha,\beta)\in E[\Pi_s]:  x\in\calX_s(\beta)}}  f^-(x) \right]
\label{eq:DP'':toProve}
\end{equation}
where $\llbracket\cdot\rrbracket=\mathds{1}$ if the argument is true,
and $\mathbb{O}$ otherwise.
This will be sufficient for establishing the theorem: summing these equations over $x\in\calX_s(\alpha)$
and using~\eqref{eq:DP'':MandW}, \eqref{eq:DP'':Wminus} yields eq.~\eqref{eq:alg'':Mupdate}.

Two cases are possible:

\noindent{\bf Case 1}: $x\in\calX_s(\beta)$ for some $(\alpha,\beta)\in E[\Pi_s]$.
(Such $\beta$ is unique since sets $\calX_s(\beta)$ are disjoint.)
Then both sides of~\eqref{eq:DP'':toProve} are $\mathbb{O}$.

\noindent{\bf Case 2}: $x\in\calX_s(\alpha;\Pi_s)$. Then eq.~\eqref{eq:DP'':toProve}
is equivalent to $f(x)=\phi(\alpha)\otimes f^-(x)$.
This holds since there is no pattern $\gamma\in\Pi^\circ_s(x)$
with $|\gamma|>|\alpha|$ (otherwise we would have $\gamma\in\Pi_s$ and thus
  $x\notin\calX_s(\alpha;\Pi_s)$ by definition~\eqref{eq:calXs'} - a contradiction).
\fi


\section{Sampling}\label{sec:sampling}
In this section consider the semiring $(R,\oplus,\otimes)=(\mathbb R,+,\times)$
from Example 1. We assume that all costs $c_\alpha$ are strictly positive.
We present an algorithm for sampling labelings $x\in D^{1:n}$ according
to the probability distribution $p(x)=f(x)/Z$.

As in the previous section, we assume that $D\subseteq\Gamma$,
and define $\Pi$ to be the set of prefixes of patterns in $\Pi^\circ$
(eq.~\eqref{eq:GLADGAKGADF}).
For a node $\alpha\in\Pi_s$ let $T_s(\alpha)$ be the
set of nodes in the subtree of $G[\Pi_s]$ rooted at $\alpha$, with
$\alpha\in T_s(\alpha)\subseteq\Pi_s$. For a pattern $\alpha\in \Pi_{s+1}-\{\varepsilon_{s+1}\}$ we define set
\begin{equation}
\Delta_{s}(\alpha)=T_{s}(\alpha^-)-\bigcup_{(\alpha,\beta)\in G[\Pi_{s+1}]}T_{s}(\beta^-)
\end{equation}
We can now present the algorithm (see Algorithm~\ref{alg:sampling}).
%
\begin{algorithm}[!h]
\caption{Sample $x\sim p(x)=f(x)/Z$}\label{alg:sampling}
\begin{algorithmic}[1]
\STATE
run Algorithm~\ref{alg:DP''} to compute messages $M_s(\alpha)$ for all patterns $\alpha=([\cdot,s],\cdot)\in\Pi$
\STATE sample $\alpha_n\!\in\!\Pi_n$ with probability $p(\alpha_n)\!\propto\! M_n(\alpha_n)$
\STATE for $s=n\!-\!1,\ldots,1$ sample $\alpha_s\in\Delta_s(\alpha_{s+1})$ with probability
$p(\alpha_s)\propto M_s(\alpha_s)$
\STATE return labeling $x$ with $x_{s:s}=(\alpha_s)_{s:s}$ for $s\in[1,n]$
\end{algorithmic}
\end{algorithm}

We say that step $s$ of the algorithm is {\em valid} if
either (i) $s=n$, or (ii) $s\in[1,n-1]$, step $s+1$ is valid,
$\alpha_{s+1}\ne\varepsilon_{s+1}$ and $M_s(\alpha)>0$
for some $\alpha\in\Delta_s(\alpha_{s+1})$.
(This is a recursive definition.) Clearly, if step $s$ is valid
then line 3 of the algorithm is well-defined.

\begin{theorem}
(a) With probability 1 all steps of the algorithm are valid.
(b) The returned labeling $x\in D^{1:n}$ is distributed
according to $p(x)=f(x)/Z$.
\label{th:sampling:correctness}
\end{theorem}

\myparagraph{Complexity} Assume that we have an oracle that produces independent samples from the uniform distribution on $[0,1]$
in $O(1)$ time.

The main subroutine performed by the algorithm
is sampling from a given discrete distribution. Clearly, this
can be done in $\Theta(N)$ time where $N$ is the number of allowed values of the random variable.
With a $\Theta(N)$ preprocessing, a sample can also be produced in $O(1)$ time by
the so-called ``alias method''~\cite{Vose:91}.

This leads to two possible complexities: (i) $\Theta(nP)$  (without preprocessing);
(ii) $\Theta(n)$ per sample (with preprocessing).
Let us discuss the complexity of this preprocessing.
Running Algorithm~\ref{alg:DP''} takes $\Theta(nP)$ time.
After that, for each $\alpha\in\Pi_{s+1}$ we need to run the linear time
procedure of~\cite{Vose:91} for distributions
$p(\beta)\propto M_s(\beta),\beta\in \Delta_s(\alpha_{s+1})$.
The following theorem implies that this takes $\Theta(nP|D|)$ time.
%
\begin{theorem}
There holds $\sum_{\alpha\in\Pi_{s}-\{\varepsilon_s\}}|\Delta_{s-1}(\alpha)|=|\Pi_{s-1}|\cdot|D|$.
\label{th:sampling:complexity}
\end{theorem}
\ifTR
\begin{proof}
Consider pattern $\beta\in\Pi_{s-1}$. For a letter $a\in D^{s:s}$
let $\beta^a\in\Pi_{s}$ be longest suffix $\alpha$ of $\beta a$ with $\alpha\in\Pi_s$
(at least one such suffix exists, namely $a$).
It can be seen that the set $\{\beta^a\:|\:a\in D^{s:s}\}$ is exactly
the set of patterns $\alpha\in\Pi_s-\{\varepsilon_s\}$ for which $\Delta_{s-1}(\alpha)$ contains $\beta$
(checking this fact is just definition chasing). Therefore, the sum in the theorem
equals $\sum_{\beta\in\Pi_{s-1}}|\{\beta^a\:|\:a\in D^{s:s}\}|=|\Pi_{s-1}|\cdot|D|$.
\end{proof}
\else
\fi
To summarize, we showed that with a $\Theta(nP|D|)$ preprocessing we can compute independent samples from $p(x)$
in $\Theta(n)$ time per sample.

\ifTR
\subsection{Proof of Theorem~\ref{th:sampling:correctness}}
Suppose that step $s\in[1,n]$ of the algorithm is valid;
this means that patterns $\alpha_t$ for $t\in[s,n]$ are well-defined.
For $t\in[s,n]$ we then define the set
of patterns $\calA_t  =  \Delta_t(\alpha_{t+1})\subseteq\Pi_t$
(if $t=n$ then we define $\calA_t  = \Pi_t$ instead).
We also define sets of labelings
\begin{eqnarray}
\calY_t(\alpha)         & = & \{y x_{t+1:n}\:|\: y \in\calX_t(\alpha;\Pi_t)\} \quad \forall \alpha\in\calA_t \quad \\
\calY_t & = & \calY_t(\alpha_t)
\end{eqnarray}
where $x$ is a labeling with $x_{t:t}=(\alpha_t)_{t:t}$ for $t\in[s,n]$.
Let $\calY_{n+1}=D^{1:n}$.
\begin{lemma}
Suppose that step $s\in[1,n]$ is valid. \\
(a) $\calY_{s+1}$ is a disjoint union of sets $\calY_s(\alpha)$ over $\alpha\in\calA_s$. \\
(b) For each $y\in\calY_{s+1}=\bigcup_{\alpha\in\calA_s}\calY_s(\alpha)$ there holds
$f(y)=const_s\cdot f(y_{1:s})$,
and consequently for any $\alpha\in\calA_s$ there holds
$$\sum_{y\in\calY_s(\alpha)}f(y)=const_s\cdot\sum_{y\in\calX_s(\alpha;\Pi_s)}f(y)=const_s\cdot M_s(\alpha)$$
\label{lemma:sampling}
\end{lemma}
Theorem~\ref{th:sampling:correctness} will follow from this lemma.
Indeed, the lemma shows that the algorithm implicitly computes a sequence
of nested sets $D^{1:n}=\calY_{n+1}\supseteq\calY_n\supseteq\ldots\supseteq\calY_1=\{x\}$.
At step $s$ we divide set $\calY_{s+1}$ into disjoint subsets $\calY_s(\alpha)$, $\alpha\in\calA_s$
and select one of them, $\calY_s=\calY_s(\alpha_s)$, with the probability
proportional to $M_s(\alpha_s)\propto\sum_{y\in\calY_s(\alpha_s)} f(y)$.

We still need to show that if step $s\in[2,n]$ is valid
then step $s-1$ is valid as well with probability 1. It follows from the precondition
that $\alpha_s$ sampled in line 3 satisfies $M_s(\alpha_s)>0$ with probability 1;
this implies that $\alpha_s\ne\varepsilon_s$.
From the paragraph above we get that $\sum_{y\in\calY_s}f(y)>0$ with probability 1.
We also have $\sum_{\alpha\in\calA_{s-1}} M_{s-1}(\alpha)\propto \sum_{\alpha\in\calA_{s-1}} \sum_{y\in\calY_{s-1}(\alpha)}f(y)=\sum_{y\in\calY_s}f(y)>0$
implying that $M_{s-1}(\alpha)>0$ for some $\alpha\in\calA_{s-1}$.
This concludes the proof that step $s-1$ is valid with probability 1.

It remains to prove Lemma \ref{lemma:sampling}.

\myparagraph{Part (a)} First, we need to check that
$\calX_{s}(\alpha_{s+1}^-;\Pi_{s+1}^{-})$ is equal to the disjoint union of $\calX_{s}(\alpha;\Pi_{s})$ over
$\alpha\in \Delta_{s}(\alpha_{s+1})$ where $\Pi_{s+1}^{-} = \left\{\alpha^-\:|\:\alpha\in \Pi_{s+1}\right\}$.
Disjointness of $\calX_{s}(\alpha;\Pi_{s})$ for different $\alpha\in\Pi_s$ is obvious. Since $\Pi_{s+1}^{-}\subseteq \Pi_{s}$,
then for any $\alpha\in \Delta_{s}(\alpha_{s+1})$, $\calX_{s}(\alpha;\Pi_{s})\subseteq \calX_{s}(\alpha_{s+1}^-;\Pi_{s+1}^{-})$ is straightforward
from the definition of $\Delta_{s}(\alpha_{s+1})$.
Thus, we only need to check the inclusion of $\calX_{s}(\alpha_{s+1}^-;\Pi_{s+1}^{-})$ in the union.

Elements of $\Pi_{s+1}^{-}\cup\left\{\alpha_{s+1}^-\right\}$ can be seen as nodes in tree $G[\Pi_{s}]$.
Then any pattern $x$ from $\calX_{s}(\alpha_{s+1}^-;\Pi_{s+1}^{-})$ defines the longest suffix $s(x)$ such that $s(x)\in \Pi_{s}$.
It is easy to see that $s(x)\in T_s(\alpha_{s+1}^-)$, and moreover, the descending path in $G[\Pi_{s}]$ from $\alpha_{s+1}^-$ to $s(x)$ does not contain elements
from $\Pi_{s+1}^{-}-\left\{\alpha_{s+1}^-\right\}$, otherwise $x, s(x)\notin \calX_{s}(\alpha_{s+1}^-;\Pi_{s+1}^{-})$. It is easy to see that
this is equivalent to $s(x)\in \Delta_{s}(\alpha_{s+1})$. Since $x\in \calX_{s}(s(x);\Pi_{s})$, $\calX_{s}(\alpha_{s+1}^-;\Pi_{s+1}^{-})$ is a subset of the
union of $\calX_{s}(\alpha;\Pi_{s})$ over $\alpha\in \Delta_{s}(\alpha_{s+1})$.

Now according to definition of $\calY_{s+1}$ we can write:
\begin{eqnarray*}
\calY_{s+1} &=& \{y x_{s+2:n}\:|\: y \in\calX_{s+1}(\alpha_{s+1};\Pi_{s+1})\}  \\
&=&\{y (\alpha_{s+1})_{s+1:s+1}x_{s+2:n}\:|\: y \in\calX_{s}(\alpha_{s+1}^-;\Pi_{s+1}^-)\}  \\
&=&\bigcup_{\alpha\in \Delta_{s}(\alpha_{s+1})}\{y (\alpha_{s+1})_{s+1:s+1}x_{s+2:n}\:|\: y \in\calX_{s}(\alpha;\Pi_{s})\}
\end{eqnarray*}
It only remains to check that in the last union the set corresponding to $\alpha\in \Delta_{s}(\alpha_{s+1})$ is exactly equal to $\calY_s(\alpha)$.

\myparagraph{Part (b)}
Let $p$ be the start position of $\alpha_{s+1}$, i.e.\ $\alpha_{s+1}=([p,s+1],\cdot)$.
Consider labeling $y\in\calY_{s+1}$, we then must have $y=\ast\alpha_{s+1}\ast$.
Let $\beta=([i,j],\cdot)$ be a pattern with $y=\ast\beta\ast$, $j>s$.
We will prove that $i\ge p$; this will imply the claim.

Suppose on the contrary that $i<p$. Denote $\gamma=\beta_{i:s+1}$,
then $\gamma\in \Pi_{s+1}$ and $\gamma=+\alpha_{s+1}$.
Therefore, $y_{1:s+1}\notin \calX_{s+1}(\alpha_{s+1};\Pi_{s+1})$ (since $y_{1:s+1}=\ast\gamma$).
However, this contradicts the assumption that $y\in\calY_{s+1}=\calY_{s+1}(\alpha_{s+1})$.
\fi


\section{Computing marginals}\label{sec:marginals}
In this section we again consider the semiring $(R,\oplus,\otimes)=(\mathbb R,+,\times)$
from Example 1 where all costs $c_\alpha$ are strictly positive,
and consider a probablity distribution $p(x)=f(x)/Z$ over labelings $x\in D^{1:n}$.

For a pattern $\alpha$ we define
\begin{eqnarray}
\Omega(\alpha)&=&\{x\in D^{1:n}\:|\:x=\ast\alpha\ast\} \\
Z(\alpha)&=&\sum_{x\in\Omega(\alpha)} f(x)\label{eq:marginals:Zij}
\end{eqnarray}
We also define the set of patterns
\begin{equation}
\Pi=\{\alpha\:|\:\exists \alpha\ast,\ast\alpha\in\Pi^\circ,\alpha\mbox{ is non-empty}\}
\end{equation}
Note that $\Pi^\circ\subseteq\Pi$ and $|\Pi_s|=|I(\Gamma)|$ for indexes $s$
that are sufficiently far from the boundary.
We will present an algorithm for computing $Z(\alpha)$ for all
patterns $\alpha\in\Pi$ in time $O(n\sum_{\alpha\in I(\Gamma)}|\alpha|)$.
Marginal probabilities
of a pattern-based CRF can then be computed as $p(x_{i:j}=\alpha)=Z(\alpha)/Z$
for a pattern $\alpha=([i,j],\cdot)$.

In the previous section we used graph $G[\Pi_s]$ for a set of patterns $\Pi_s$;
here we will need an analogous but a slightly different construction for patterns in $\Pi$.
For patterns $\alpha,\beta$ we write $\alpha\sqsubseteq \beta$
if $\beta=\ast\alpha\ast$.
If we have $\beta=+\alpha+$
then we write $\alpha\sqsubset \beta$.

Now consider $\alpha\in\Pi$. We define $\Phi(\alpha)$ to be the set of patterns $\beta\in\Pi$
such that $\alpha\sqsubset \beta$ and there is no other
pattern
 $\gamma\in\Pi$ with $\alpha\sqsubset \gamma\sqsubseteq\beta$.

Our algorithm is given below. In the first step it runs Algorithm~\ref{alg:DP''}
from left to right and from right to left; as a result, we get
forward messages $\overrightarrow{W}_j(\alpha)$
and backward messages $\overleftarrow{W}_i(\alpha)$
for patterns $\alpha=([i,j],\cdot)$ such that
\begin{equation}
\overrightarrow{W}_j(\alpha)=\sum_{\substack{x=\ast\alpha \\ x\in D^{1:j}}} f(x)\qquad
\overleftarrow{W}_i(\alpha)=\sum_{\substack{y=\alpha\ast \\ y\in D^{i:n}}}f(y)
\label{eq:marginals:WFwBw}
\end{equation}

\begin{algorithm}
\caption{Computing values $Z(\alpha)$}\label{alg:marginals}
\begin{algorithmic}[1]
\STATE
run Algorithm~\ref{alg:DP''} in both directions to get messages $\overrightarrow{W}_j(\alpha)$, $\overleftarrow{W}_i(\alpha)$.
For each pattern $\alpha=([i,j],\cdot)\in\Pi$ set
\begin{subequations}
\label{eq:marginals:stepOne}
\begin{eqnarray}
W(\alpha)  &:=&\frac{\overrightarrow{W}_j(\alpha) \overleftarrow{W}_i(\alpha)}{f(\alpha)} \label{eq:marginals:Wij} \\
W^-(\alpha)&:=&\frac{\overrightarrow{W}_{j-1}(\alpha_{i:j-1})
\overleftarrow{W}_{i+1}(\alpha_{i+1:j})}{f(\alpha_{i+1:j-1})} \label{eq:marginals:WijMinus}
\end{eqnarray}
\end{subequations}
\STATE for  $\alpha\!\in\!\Pi$ (in the order of decreasing $|\alpha|$)  set
\begin{equation}
 Z(\alpha) :=W(\alpha)
+
\sum_{\beta\in\Phi(\alpha)}
 \left[Z(\beta) - W^-(\beta)\right]
\label{eq:alg:marginals:update}
\end{equation}
\end{algorithmic}
\end{algorithm}

\ifTR
\begin{theorem}
Algorithm~\ref{alg:marginals} is correct.
\label{th:marginals}
\end{theorem}
We prove the theorem in section~\ref{sec:marginals:proof}, but first let us discuss algorithm's complexity.
\fi
We claim that all values $f(\alpha)$ used by the algorithm can be computed in $O(n(\PallRight+\PallLeft))$ time
where $\PallRight$ and $\PallLeft$ are respectively the number of distinct non-empty prefixes and suffixes of words in $\Gamma$.
Indeed, we first compute these values for patterns in the set $\overrightarrow\Pi\triangleq\{\alpha\:|\:\exists\alpha\ast\in\Pi^\circ\}$;
by Lemma~\ref{lemma:varphis}, this takes $O(n\PallRight)$ time. This covers values $f(\alpha)$ used in eq.~\eqref{eq:marginals:Wij}.
As for the value in eq.~\eqref{eq:marginals:WijMinus} for pattern $\alpha=([i,j],\cdot)\in\Pi$, we can use the formula
$$
f(\alpha_{i+1:j-1})=
\frac
{f(\alpha)\tilde c_\alpha}
{\overrightarrow\phi(\alpha)\overleftarrow\phi(\alpha)}
$$
where $\tilde c_\alpha=c_\alpha$ if $\alpha\in\Pi^\circ$ and $\tilde c_\alpha=1$ otherwise, and
\begin{eqnarray*}
\overrightarrow\phi(\alpha)=\hspace{-10pt}\prod_{\beta\in\Pi^\circ,\alpha=\ast\beta}\hspace{-10pt}c_\beta\;,&&
\qquad\overleftarrow\phi(\alpha)=\hspace{-10pt}\prod_{\beta\in\Pi^\circ,\alpha=\beta^\ast}\hspace{-10pt}c_\beta
\end{eqnarray*}
The latter values can be computed in $O(n( \PallRight+\PallLeft))$ time by applying Lemma~\ref{lemma:varphis}
in the forward and backward directions. (In fact, they were already computed when running Algorithm \ref{alg:DP''}.)

We showed that step 1 can be implemented in $O(n( \PallRight+\PallLeft))$ time; let us analyze step 2.
The following lemma implies that it performs $O(n\sum_{\alpha\in I(\Gamma)}|\alpha|)$ arithmetic operations;
since $\sum_{\alpha\in I(\Gamma)}|\alpha|\ge \sum_{\alpha\in \Gamma}|\alpha|\ge \max\left\{\PallRight, \PallLeft  \right\}$, we then get that the
overall complexity is $O(n\sum_{\alpha\in I(\Gamma)}|\alpha|)$.
\begin{lemma}
For each $\beta\in\Pi$ there exist at most $2|\beta|$ patterns
$\alpha\in\Pi$ such that $\beta\in\Phi(\alpha)$.
\end{lemma}
\ifTR
\begin{proof}
Let $\Psi$ be the set of such patterns $\alpha$. 
Note, there holds $\beta=+\alpha+$.
We need to show that $m\triangleq|\Psi|\le 2|\beta|$.
Let us order patterns $\alpha=([i,j],\cdot)\in\Psi$ lexicographically (first by $i$, then by $j$):
$\Psi=\{\alpha_1,\ldots,\alpha_m\}$ with $\alpha_t=([i_t,j_t],\cdot)$, and denote $\sigma_t=(i_t-k)+(j_t-k)\in[2,2(\ell-k-1)]$
where $[k,\ell]$ is the interval for $\beta$.
We will prove by induction that $\sigma_t\ge t+1$ for $t\in[1,m]$;
this will imply that $m+1\le \sigma_m\le 2(\ell-k-1)=2(|\beta|-2)$, as desired.

The base case is trivial. Suppose that it holds for $t-1$; let us prove it for $t$.
If $i_t=i_{t-1}$ then $j_t>j_{t-1}$ by the definition of the order on $\Psi$,
so the claim holds. Suppose that $i_t>i_{t-1}$. If $j_t<j_{t-1}$
then $\alpha_t\sqsubset \alpha_{t-1} \sqsubset \beta$
contradicting the condition $\beta \in\Phi(\alpha_t)$.
Thus, $j_t\ge j_{t-1}$, and so the claim of the induction step holds.
\end{proof}
\else
\fi

\myparagraph{Remark 1} { An alternative method for computing marginals
with complexity $O\left(n |\Gamma| L^2 \ell_{\max}^2\right)$
was given in~\cite{Ye:NIPS09}.
They compute value $Z(\alpha)$ directly from messages $\overrightarrow{M}_{j'}(\cdot)$ and $\overleftarrow{M}_{i'}(\cdot)$
by summing over {\bf pairs} of patterns (thus the square factor in the complexity).
In contrast, we use a recursive rule that uses previously computed values of $Z(\cdot)$.
We also use the existence of the ``$\ominus$'' operation.
This allows us to achieve better complexity.
}

\ifTR
\subsection{Proof of theorem~\ref{th:marginals}}\label{sec:marginals:proof}
Consider labeling $x\in D^{1:n}$. We define $\Lambda(x)=\{\alpha\in\Pi^\circ\:|\:x=\ast\alpha\ast\}$
to be the set of patterns contained in $x$.
For an interval $[i,j]\subseteq[1,n]$ we also define
 sets
\begin{subequations}
\begin{eqnarray}
\Lambda_{ij}(x)  &=&\{\beta\in\Lambda(x) \:|\: \beta=+x_{i:j}+\} \\
\Lambda^-_{ij}(x)&=&\{\beta\in\Lambda(x) \:|\: \beta=\ast x_{i:j} \ast \}
\end{eqnarray}
\end{subequations}
and corresponding costs
\begin{subequations}
\begin{eqnarray}
f_{ij}(x)&=&\prod_{\beta\in\Lambda(x)-\Lambda_{ij}(x)}c_\beta \\
f^-_{ij}(x)&=&\prod_{\beta\in\Lambda(x)-\Lambda^-_{ij}(x)}c_\beta
\end{eqnarray}
\end{subequations}
It can be checked that quantities $W(\alpha)$, $W^-(\alpha)$ defined via~\eqref{eq:marginals:WFwBw}
and~\eqref{eq:marginals:stepOne}
satisfy
\begin{equation}
W(\alpha)  =\hspace{-5pt}\sum_{x\in\Omega(\alpha)} \hspace{-5pt}f_{ij}(x) \qquad
W^-(\alpha)=\hspace{-5pt}\sum_{x\in\Omega(\alpha)} \hspace{-5pt}f^-_{ij}(x)
\label{eq:marginals:Wij'}
\end{equation}
where $[i,j]$ is the interval for $\alpha$.

Consider pattern $\alpha=([i,j],\cdot)\in\Pi$.
We will show that for any $x\in\Omega(\alpha)$ there holds
\begin{equation}
f(x) =f_{ij}(x)
+
\sum_{\substack{\beta=([k,\ell],\cdot)\in\Phi(\alpha):  x=\ast\beta\ast}}
 \left[f(x) - f^-_{k\ell}(x)\right]
\label{eq:marginals:toProve}
\end{equation}
This will be sufficient for establishing algorithm's correctness:
summing these equations over $x\in\Omega(\alpha)$ and using~\eqref{eq:marginals:Zij},\eqref{eq:marginals:Wij'}
 yields eq.~\eqref{eq:alg:marginals:update}.

\begin{lemma}
The sum in~\eqref{eq:marginals:toProve} contains at most one pattern $\beta=([k,\ell],\cdot)\in\Phi(\alpha)$
with $x=\ast \beta\ast$.
\end{lemma}
\begin{proof}
Consider two such patterns $\beta^1=([k^1,\ell^1],\cdot)$ and $\beta^2=([k^2,\ell^2],\cdot)$.
Define $k\!=\!\max\{k^1,k^2\}$, $\ell\!=\!\min\{\ell^1,\ell^2\}$, $\beta\!=\!x_{k:\ell}$, then $\alpha\!\sqsubset\!\beta\!\sqsubseteq\!\beta^t$
for $t\!\in\!\{1,2\}$.
Using the definition of set $\Pi$, it can be checked that $\beta\in\Pi$.
The fact that $\beta^t\in\Phi(\alpha)$ then implies that
$\beta^t=\beta$ for $t\in\{1,2\}$, and so $\beta^1=\beta^2$.
\end{proof}
We now consider two possible cases.

\noindent{\bf Case 1}: There are no patterns $\beta=([k,\ell],\cdot)\in\Phi(\alpha)$ with
$x=\ast\beta\ast$. This implies that $\Lambda_{ij}(x)$ is empty, and therefore $f(x)=f_{ij}(x)$.
Eq.~\eqref{eq:marginals:toProve} thus holds.

\noindent{\bf Case 2}: There exists a (unique) pattern $\beta=([k,\ell],\cdot)\in\Phi(\alpha)$ with
$x=\ast\beta\ast$. Eq.~\eqref{eq:marginals:toProve} then becomes equivalent to
the condition $f_{ij}(x)=f^-_{k\ell}(x)$. We will prove this by showing that  $\Lambda_{ij}(x)=\Lambda^-_{k\ell}(x)$.

The inclusion $\Lambda_{k\ell}^-(x)\subseteq \Lambda_{ij}(x)$ is obvious; let us show the other direction.
Suppose that $\gamma=([p,q],\cdot)\in\Lambda_{ij}(x)$.
Define $\hat p=\max\{k,p\}$, $\hat q=\min\{q,\ell\}$, $\hat\gamma=x_{\hat p:\hat q}$.
It can be  checked that $\hat\gamma\in\Pi$.
We also have $\alpha\sqsubset \hat\gamma \sqsubseteq \beta$.
Therefore, condition $\beta\in\Phi(\alpha)$ implies that $\hat\gamma=\beta$,
and so $p\le k$, $q\ge \ell$, and $\gamma\in\Lambda^-_{k\ell}(x)$.
\else
\fi


\section{General case: $O(n\Pproper|D|)$ algorithm}\label{sec:semiring:D}
In this section and in the next one we consider the case of a general commutative semiring $(R,\oplus,\otimes)$
(without assuming the existence of an inverse operation for $\oplus$).
This can be used for computing MAP in CRFs containing positive costs $c_\alpha$.
The algorithm closely resembles the method in \cite{Ye:NIPS09}; it is based on the same idea
and has the same complexity. Our primary goal of presenting this algorithm is to motivate
the $O(n\Pall\log(\ell_{\max}+1))$ algorithm for the MAP problem given in the next section.

First, we select $\Pi$ as
the set of proper prefixes of patterns in $\Pi^\circ$:
\begin{equation}
\Pi         =\{\alpha\:|\:\exists\alpha+\in\Pi^\circ\}
\end{equation}
For each $\alpha\in\Pi_s$ we will compute message
\begin{equation}
M_s(\alpha)=\bigoplus_{x\in\calX_s(\alpha;\Pi_s)} f(x)
\label{eq:general:Ms}
\end{equation}
In order to go from step $s\!-\!1$ to $s$, we will use an extended set of patterns $\widehat\Pi_s$:
\begin{eqnarray}
\widehat \Pi_s&=&\{\alpha\:|\:\alpha^-\in\Pi_{s-1}\}\cup\{\varepsilon_s\}  \\
&=&\{\alpha c\:|\:\alpha\in \Pi_{s-1},c\in D^{s:s}\}\cup\{\varepsilon_s\} \nonumber
\label{eq:general:widehatPis}
\end{eqnarray}
It can be checked that
\begin{equation}
\Pi_s\subseteq\widehat\Pi_s\mbox{~~and~~}\Pi^\circ_s\subseteq\widehat\Pi_s
\label{eq:widehatPi}
\end{equation}
In step $s$ we compute values $M_s(\alpha)$ in eq.~\eqref{eq:general:Ms} for all $\alpha\in\widehat\Pi_s$.
Note, we now use the generalized definition of $\calX_s(\alpha;\Pi_s)$
(eq.~\eqref{eq:calXs'})
since we may have $\alpha\notin\Pi_s$.
After completing step $s$, messages $M_s(\alpha)$
for $\alpha\in\widehat\Pi_s\!-\!\Pi_s$ can be discarded.

Our algorithm is given below. We have $|\widehat\Pi_s|\!=\! \Pproper|D|+1\!$
for indexes $s$ that are sufficiently far from the boundary,
and thus the algorithm's complexity is $\Theta(n\Pproper|D|)$
(if Lemma~\ref{lemma:varphis} is used for computing values $\phi(\alpha)$).

\begin{algorithm}
\caption{Computing $Z=\bigoplus_{x\in D^{1:n}} f(x)$}\label{alg:DP'}
\begin{algorithmic}[1]
\STATE initialize messages:
set $M_0(\varepsilon_0):=\mathbb{O}$
\STATE for  $s=1,\ldots,n$ traverse nodes $\alpha\in \widehat\Pi_s$ of tree $G[\widehat\Pi_s]$
starting from the leaves and set
\begin{equation}
M_{s}( \alpha ) := \left[\phi(\alpha)\otimes M_{s-1}(\alpha^-)\right] \oplus
  \bigoplus_{(\alpha,\beta)\in E[\widehat\Pi_s],\beta\notin \Pi_s} M_s(\beta)
\label{eq:message:update}
\end{equation}
If $\alpha=\varepsilon_s$ then use $M_{s-1}(\alpha^-)=\mathbb{O}$
\STATE return
$
Z:=\bigoplus_{\alpha\in\Pi_n}M_n(\alpha)
$
\end{algorithmic}
\end{algorithm}

\begin{theorem} Algorithm~\ref{alg:DP'} is correct.
\label{th:correctness'}
\end{theorem}

\myparagraph{Remark 2} {
As we already mentioned, Algorithm~\ref{alg:DP'} resembles the algorithm in~\cite{Ye:NIPS09}.
The latter computes the same set of messages as we do 
but using the following recursion: for a pattern $\alpha\in \Pi_s-\{\varepsilon_s\}$
they set
\begin{eqnarray}
M_s(\alpha):=
\bigoplus_{\gamma\in T_{s-1}(\alpha^-)-\bigcup\limits_{(\alpha,\beta)\in E[\Pi_s]}T_{s-1}(\beta^-)}
\phi(\gamma a)\otimes M_{s-1}(\gamma)
\label{eq:YeUpdate}
\end{eqnarray}
where $a=\alpha_{s:s}$ is the last letter of $\alpha$ and $T_{s-1}(\beta)=\{\gamma\:|\:\gamma=\ast\beta,\gamma\in\Pi_{s-1}\}$ for $\beta\in\Pi_{s-1}$ is
the set of patterns in the branch of $G[\Pi_{s-1}]$ rooted at $\beta$.
It can be shown that updates~\eqref{eq:message:update} and~\eqref{eq:YeUpdate} are equivalent:
they need exactly the same number of additions (and the same number of
multiplications, if $\phi(\gamma a)$ in eq.~\eqref{eq:YeUpdate} is replaced with $\phi(\alpha)$ and moved before the sum).
%
}
\ifTR
\subsection{Proof of Theorem \ref{th:correctness'}}
To prove the correctness,
we need to show that eq.~\eqref{eq:message:update} holds for each $\alpha\in \widehat\Pi_s$.

\begin{lemma}
For any $\alpha\in\widehat\Pi_s$ there holds
\begin{equation}
\phi(\alpha)\otimes M_{s-1}(\alpha^-)=\bigoplus_{x\in\calX_s(\alpha;\widehat \Pi_s)} f(x)
\label{eq:general:widehatM}
\end{equation}
\label{lemma:general:widehatM}
\end{lemma}
\begin{proof}
For $\alpha=\varepsilon_s$ the claim is trivial:
we have $D^{s:s}\subseteq\widehat\Pi_s$ (since $\varepsilon_{s-1}\in\Pi_{s-1}$), therefore
$\calX_s(\alpha;\widehat \Pi_s)=\varnothing$ and the sum in~\eqref{eq:general:widehatM} is $\mathbb{O}$.
We thus assume that $\alpha\in\widehat\Pi_s-\{\varepsilon_s\}$.
Using definition~\eqref{eq:general:widehatPis}, it can be checked that
the mapping $x\mapsto x^-$ is a bijection
$\calX_s(\alpha;\widehat \Pi_s)\rightarrow \calX_{s-1}(\alpha^-;\Pi_{s-1})$.
Consider $x\in \calX_s(\alpha;\widehat \Pi_s)$.
We claim that if $x=\ast\gamma$ and $\gamma\in\Pi^\circ$ then $|\gamma|\le |\alpha|$.
Indeed, we have $\gamma\in\widehat\Pi_s$ (since $\Pi^\circ_s\subseteq\widehat\Pi_s$),
and so if $|\gamma|>|\alpha|$ then $x\notin\calX_s(\alpha;\widehat \Pi_s)$ - a contradiction.

Using the claim, we conclude that $\phi(\alpha)\otimes f(x^-)=f(x)$.
This implies the lemma.
\end{proof}

The fact $\Pi_s\subseteq\widehat\Pi_s$ implies the
following characterization of $\calX_s(\alpha;{\Pi_s})$ for $\alpha\in\widehat\Pi_s$:
\begin{eqnarray*}
\calX_s(\alpha;\Pi_s)=\left\{x\in\calX_s(\alpha)
\begin{picture}(1,0)
  \put(3,-10){\line(0,1){24}}
\end{picture}
\begin{tabular}{l}
$x\ne\ast\beta$ for any $\beta$ in the subtree \\ 
of $\alpha$ in $G[\widehat\Pi_s]$ with $\beta\in\Pi_s$, $\beta\ne\alpha$
\end{tabular}
\hspace{-3pt}\right\}
\end{eqnarray*}
where the subtree of $\alpha$ in $G[\widehat\Pi_s]$ is defined as the set of
descendants of $\alpha$ in $G[\widehat\Pi_s]$ (including $\alpha$).

Now it becomes clear that $\calX_s(\alpha;{\widehat \Pi_s})\subseteq \calX_s(\alpha;{\Pi_s})$ and $\calX_s(\alpha;{\Pi_s})-\calX_s(\alpha;{\widehat \Pi_s})$ equals the set of partial labelings $x\in\calX_s(\alpha)$ such that
\begin{itemize}
\item $x$ ends with $\beta\in \widehat\Pi_s-\Pi_s$, $(\alpha,\beta)\in E[\widehat\Pi_s]$, and
\item $x$ does not end with any pattern $\gamma\in\Pi_s$ from the subtree of $\beta$ in $G[\widehat\Pi_s]$.
\end{itemize}
It is easy to check that the last set of partial labelings  equals $\calX_s(\beta;{\Pi_s})$,
and such sets for different $\beta$'s are disjoint.
We showed that $\calX_s(\alpha;{\Pi_s})$ is a disjoint union of sets $\calX_s(\alpha;{\widehat \Pi_s})$
and $\calX_s(\beta;{\Pi_s})$ for $(\alpha,\beta)\in E[\widehat\Pi_s], \beta\notin \Pi_s$.
This fact together with Lemma~\ref{lemma:general:widehatM}
implies eq.~\eqref{eq:message:update}.
\else
\fi


\section{General case: $O(n\Pall\log \Pall)$ algorithm}\label{sec:semiring:logL}
In the previous section we presented an algorithm for a general commutative semiring
with complexity $O(n\Pproper|D|)$. In some applications the size of the input alphabet
can be very large (e.g. hundreds or thousands), so the technique may be very costly.
Below we present a more complicated version with complexity $O(n\Pall\log \Pall)$.
If $(R,\oplus,\otimes)\!=\!(\overline{\mathbb R},\min,+)$ then this can be reduced to $O(n\Pall\log (\ell_{\max}+1))$ using the algorithm for {\em Range Minimum Queries}
by~\cite{BerkmanVishkin:93}.
We assume that $D\subseteq\Gamma$.

We will use the same definitions of sets $\Pi_s$ and $\widehat\Pi_s$
as in the previous section, and the same intepretation of messages $M_s(\alpha)$
given by eq.~\eqref{eq:general:Ms}.
  We need to solve the following problem: given messages
$M_{s-1}(\alpha)$ for $\alpha\in\Pi_{s-1}$, compute messages $M_s(\alpha)$ for $\alpha\in\Pi_s$.

Recall that in the previous section this was done by computing messages $M_s(\alpha)$
for patterns in the extended set $\widehat\Pi_s$ of size $O(\Pproper|D|)$.
The idea of our modification is to compute these messages only for
patterns in the set $\Sigma_s$ where
$$
\Sigma^\circ_s\subseteq\Sigma_s\subseteq\widehat\Pi_s\qquad
\Sigma^\circ_s=\Pi_s\cup\Pi^\circ_s\qquad
|\Sigma_s|\le 2|\Sigma^\circ_s|
$$
Note that $|\Sigma^\circ_s|\le \Pall+1$.
Patterns in $\Sigma_s$ will be called {\em special}. To define them,
we will use the following notation for a node $\alpha\in\widehat\Pi_s$:
\begin{itemize}
\item $\widehat\Phi_s(\alpha)$ is the set of children of $\alpha$ in the tree $G[\widehat\Pi_s]$.
\item $\widehat T_s(\alpha)$
 is the set of nodes in the subtree of $G[\widehat\Pi_s]$ rooted at $\alpha$.
We have $\alpha\in \widehat T_s(\alpha)\subseteq\widehat\Pi_s$.
\end{itemize}
We now define set $\Sigma_s$ as follows: pattern $\alpha\in\widehat\Pi_s$ is special if either (i) $\alpha\in\Sigma^\circ_s$, or
(ii) $\alpha$ has at least two children $\beta_1,\beta_2\in\widehat\Phi_s(\alpha)$
such that subtree $\widehat T_s(\beta_i)$ for $i\in\{1,2\}$ contains a pattern from $\Sigma^\circ_s$,
i.e.\ $\widehat T_s(\beta_i)\cap\Sigma^\circ_s\ne\varnothing$.

The set of remaining patterns $\widehat\Pi_s-\Sigma_s$ will be split
into two sets $\calA_s$ and $\calB_s$ as follows:
\begin{itemize}
\item $\calA_s$ is the set of patterns $\alpha\in\widehat\Pi_s-\Sigma^\circ_s$
such that subtree $\widehat T_s(\alpha)$ does not contain patterns from $\Sigma^\circ_s$.
\item $\calB_s$ is the set of patterns $\alpha\in\widehat\Pi_s-\Sigma^\circ_s$
such that $\alpha$ has exactly one child $\beta$ in $G[\widehat\Pi_s]$
for which $\widehat T_s(\beta)\cap\Sigma^\circ_s\!\ne\!\varnothing$.
\end{itemize}
Clearly, $\widehat\Pi_s$ is a disjoint union of $\calA_s$, $\calB_s$ and $\Sigma_s$.

\ignore{
In order to define them,
we need to introduce the following notation for a pattern  $\alpha\in\Pi_s$:
\begin{itemize}\setlength{\itemsep}{-2pt}
\item $\widehat T_s(\alpha)$ is the set of patterns in the subtree of $G[\widehat\Pi_s]$
rooted at $\alpha$. Note, $\alpha\in\widehat T_s(\alpha)\subseteq\widehat\Pi_s$.
\item $T_s(\alpha)=\widehat T_s(\alpha)\cap\Sigma^\circ_s$.
\item $\widehat\Phi_s(\alpha)$ is the set of children of $\alpha$ in $G[\widehat\Pi_s]$.
\end{itemize}
Now define sets
\begin{eqnarray*}
\calA_s &\hspace{-10pt}=\hspace{-10pt}& \{\alpha\in\widehat\Pi_s\:|\:T_s(\alpha)=\varnothing\} \qquad    \overline \calA_s \;=\; \widehat\Pi_s-\calA_s \\
\calB_s &\hspace{-10pt}=\hspace{-10pt}& \{\alpha\in\widehat\Pi_s-\Sigma^\circ_s\:|\:|\widehat\Gamma(\alpha)\cap\overline \calA_s(\alpha)|=1\} \\
\Sigma_s &\hspace{-10pt}=\hspace{-10pt}& \{\alpha\in\widehat\Pi_s-\Sigma^\circ_s\:|\:|\widehat\Gamma(\alpha)\cap\overline \calA_s(\alpha)|\ge 2\}\cup\Sigma^\circ_s \\
\end{eqnarray*}
Thus, pattern $\alpha\in\widehat\Pi_s$ is special if either $\alpha\in\Pi_s$
or $\alpha$ has at least two children $\beta$ with $T_s(\beta)\ne\varnothing$.
}

\begin{figure}[t]
\vskip 0.2in
\small
\begin{center}
\begin{tabular}{c}
\includegraphics[scale=0.25]{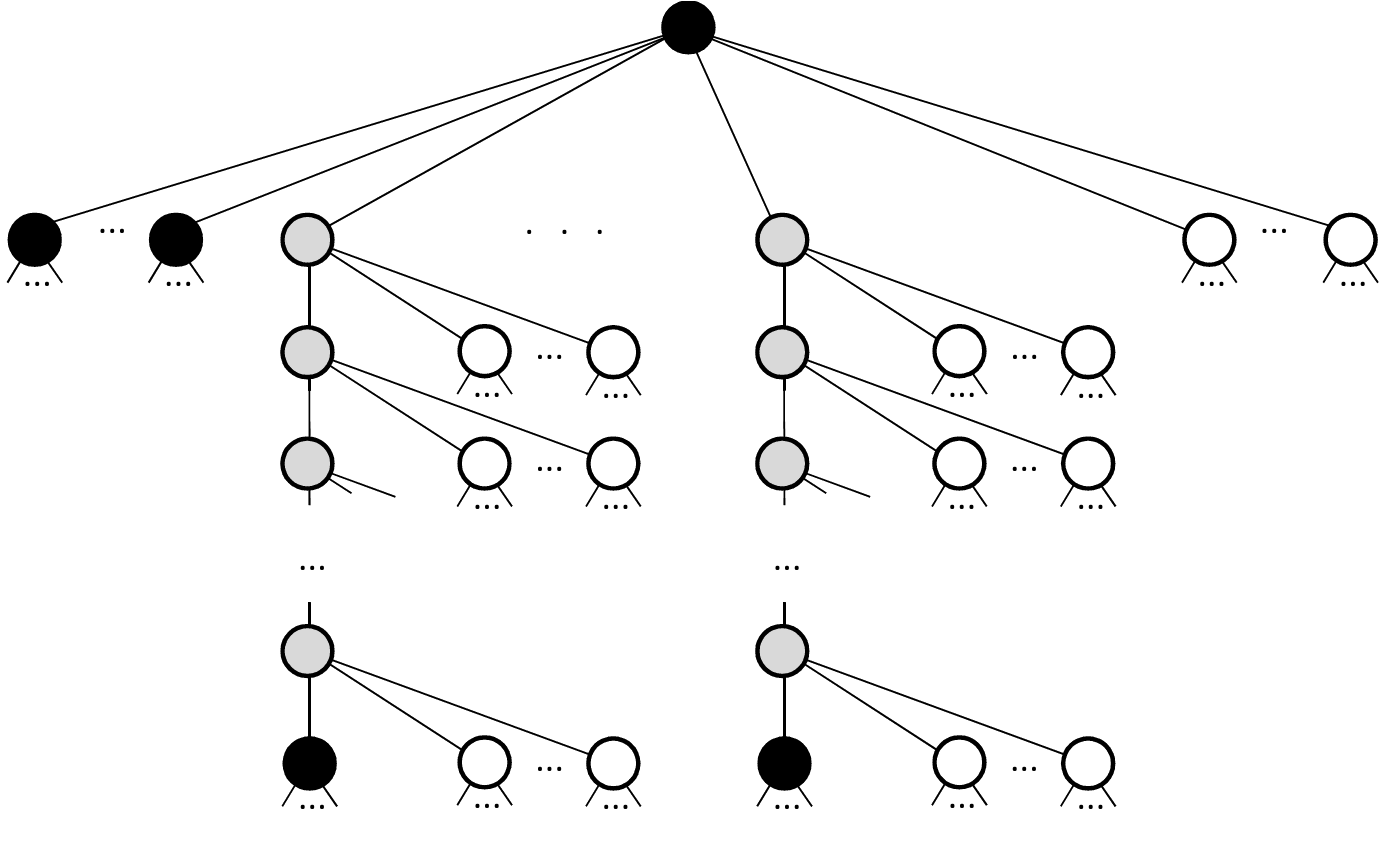} \vspace{-11pt} \\
\begin{picture}(1,0)
  \put(-5,102){$\alpha$}
  \put(-57,69){$\beta$}
  \put(-61,6){$\beta_\downarrow$}
\end{picture}
\end{tabular}
\caption{Structure of the subtree of $G[\widehat\Pi_s]$ rooted at a node $\alpha\in\Sigma_s$.
White circles represent nodes in $\calA_s$ (so all their children are also white), gray circles - nodes in $\calB_s$,
and black circles - nodes in $\Sigma_s$. Note, if $\beta\in\calB_s$ is a child of $\alpha$ then $(\alpha,\beta_\downarrow)\in E[\Sigma_s]$.
}
\label{fig:Sigmas}
\end{center}
\vskip -0.2in
\end{figure}
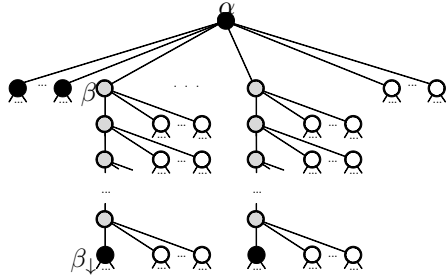

Consider a node $\alpha\in \calB_s$. From the definition, $\alpha$ has exactly one link
to a child in $G[\widehat\Pi_s]$
that belongs to $\calB_s\cup\Sigma_s$. If this child does not belong to $\Sigma_s$, then it belongs to $\calB_s$ and the same argument can be repeated
for it. By following such links we eventually
get to a node in $\Sigma_s$; the first such node will be denoted as $\alpha_\downarrow$.



We will need two more definitions. For an index $t$ and patterns $\alpha,\beta$ ending at position $t$ with $\beta=+\alpha$ we denote
\begin{subequations}\label{eq:LlogL:WV}
\begin{eqnarray}
W_t(\alpha)&=&\bigoplus_{x\in\calX_t(\alpha)} f(x) \label{eq:LlogL:W} \\
V_t(\alpha,\beta)&=&\bigoplus_{x\in\calX_t(\alpha)-\calX_t(\beta)} f(x) \label{eq:LlogL:V}
\end{eqnarray}
\end{subequations}
We can now formulate the structure of the algorithm (see Algorithm~\ref{alg:LlogL}).

\begin{algorithm}
\caption{Computing $Z=\bigoplus_{x\in D^{1:n}} f(x)$}\label{alg:LlogL}
\begin{algorithmic}[1]
\STATE initialize messages:
set $M_0(\varepsilon):=\mathbb{O}$
\STATE for each $s=1,\ldots,n$ traverse nodes $\alpha\in \Sigma_s$ of tree $G[\Sigma_s]$
starting from the leaves and set
\begin{eqnarray}
M_{s}( \alpha ) &:=& \phi(\alpha)\otimes[ M_{s-1}(\alpha^-)\oplus A_s(\alpha) \oplus B_s(\alpha) ]  \oplus
  \bigoplus_{(\alpha,\beta)\in E[\Sigma_s],\beta\notin\Pi_s} M_s(\beta)
\label{eq:message:update_special}
\end{eqnarray}
where
\begin{subequations}
\begin{eqnarray}
 A_s(\alpha)&=&\bigoplus_{\beta\in\widehat\Phi_s(\alpha)\cap\calA_s} W_{s-1}(\beta^-) \label{eq:LlogL:A} \\
 B_s(\alpha)&=&\bigoplus_{\beta\in\widehat\Phi_s(\alpha)\cap \calB_s} V_{s-1}(\beta^-,{\beta_\downarrow}^-) \label{eq:LlogL:B}
\end{eqnarray}
\end{subequations}
If $\alpha=\varepsilon_s$ then use $M_{s-1}(\alpha^-):=\mathbb{O}$
\STATE return
$
Z:=\bigoplus_{\alpha\in\Pi_n}M_n(\alpha)
$
\end{algorithmic}
\end{algorithm}

To fully specify the algorithm, we still need to describe how we compute quantities $A_s(\alpha)$
and $B_s(\alpha)$ defined by eq.~\eqref{eq:LlogL:A} and~\eqref{eq:LlogL:B}.
This is addressed by the theorem below.

\begin{theorem}
(a) Algorithm~\ref{alg:LlogL} is correct. \\
(b) There holds $|\Sigma_s|\le 2 |\Sigma^\circ_s|-1 \le 2\Pall+1$. \\
(c) Let $h$ be the maximum depth of tree $G[\Pi_{s-1}]$.
(Note, $h\le \ell_{\max}+1$.)
With an $O(\Pproper\log h)$ preprocessing, values $V_{s-1}(\alpha,\beta)$
for any $\alpha,\beta\in \Pi_{s-1}$ with $\beta=+\alpha$ can be computed in $O(\log h)$ time. \\
(d) Values $A_s(\alpha)$ for all $\alpha\in\Sigma_s$ can be computed in $O(\Pall\log \Pall)$ time,
or in $O(\Pall)$ time when $(R,\oplus,\otimes)\!=\!(\overline{\mathbb R},\min,+)$.
\label{th:LlogL}
\end{theorem}
Clearly, the theorem implies that the algorithm can be implemented in $O(n\Pall\log \Pall)$ time,
or in $O(n\Pall\log (\ell_{\max}+1))$ time when $(R,\oplus,\otimes)\!=\!(\overline{\mathbb R},\min,+)$.
To see this, observe that the sum in~\eqref{eq:LlogL:B} is effectively
over a subset of children of $\alpha$ in the tree $G[\Sigma_s]$ (see Fig.~\ref{fig:Sigmas}),
and this tree has size $O(\Pall)$.
%

\ifTR
Before presenting the proof, let us make a few remarks.
To give some insights into how we prove the theorem, let us make a few remarks.

It can be easily checked that graph $G[\widehat\Pi_s]$
has the following structure: the root $\varepsilon_s$ has $|D|$ children,
and for each child $c\in D^{s:s}\subseteq\widehat\Pi_s$ the subtree
of $G[\widehat\Pi_s]$ rooted at $c$ is isomorphismic to the tree $G[\Pi_{s-1}]$.
The isomorphism $\widehat T_s(c)\rightarrow \Pi_{s-1}$ is given by the
the mapping $\alpha\mapsto \alpha^-$.

For nodes $\alpha,\beta\in\Pi_{s-1}$ with $\beta=+\alpha$
we denote $\calP_{s-1}(\alpha,\beta)$
to be the unique path from $\alpha$ to $\beta$ in $G[\Pi_{s-1}]$
(treated as a set of edges in $E[\Pi_{s-1}]$).
Analogously, for nodes $\alpha,\beta\in\widehat\Pi_{s}$ with $\beta=+\alpha$ let
$\widehat\calP_s(\alpha,\beta)$ be the unique path from $\alpha$ to $\beta$ in $G[\widehat\Pi_s]$.
It follows from the previous paragraph that if $\alpha\ne\varepsilon_s$
then path $\widehat\calP_s(\alpha,\beta)$ is isomorphic to the path $\calP_{s-1}(\alpha^-,\beta^-)$.

\fi

\ifTR
\subsection{Proof of Theorem~\ref{th:LlogL}(a)}
The statement is equivalent to the correctness of \eqref{eq:message:update_special}.
Let us first divide the sum over $\beta$ in the last expression \eqref{eq:message:update_special} into
two parts: nodes $\beta$ that belong to $\widehat\Phi(\alpha)$ and those that do not:
\begin{eqnarray}
&&\hspace{-25pt}\bigoplus_{\beta\in\widehat\Phi_s(\alpha)\cap(\Sigma_s-\Pi_s)} M_s(\beta)
\oplus  \bigoplus_{(\alpha,\beta)\in E[\Sigma_s],\beta\notin\Pi_s\cup \Phi_s(\alpha)}  M_s(\beta)
\end{eqnarray}

It is easy to see that the second sum can be written as $\sum_{\beta\in\widehat\Phi_s(\alpha)\cap\calB_s}\overline M_s(\beta_\downarrow)$
(see Fig.~\ref{fig:Sigmas}), where
\begin{equation}
\overline{M}_s(\gamma) = \begin{cases}
M_s(\gamma) & \mbox{if }\gamma\notin\Pi_s \\
\mathbb{O} & \mbox{otherwise}
\end{cases}
\end{equation}

Using the distributive law, we can rewrite \eqref{eq:message:update_special} as
\begin{eqnarray}
&&\hspace{-25pt}M_{s}( \alpha ) := \left[\phi(\alpha)\otimes M_{s-1}(\alpha^-)\right]
\nonumber  \\
&&\hspace{10pt}\oplus \hspace{4pt} \bigoplus_{\beta\in\widehat\Phi_s(\alpha)\cap\calA_s}  \![\phi(\alpha)\otimes W_{s-1}(\beta^-)  ] \nonumber \\
&&\hspace{10pt}\oplus \hspace{4pt} \bigoplus_{\beta\in\widehat\Phi_s(\alpha)\cap \calB_s} [\phi(\alpha)\otimes V_{s-1}(\beta^-,{\beta_\downarrow}^-)]\oplus \overline{M}_s(\beta_\downarrow) \nonumber \\
&&\hspace{10pt}\oplus \hspace{0pt} \bigoplus_{\beta\in\widehat\Phi_s(\alpha)\cap(\Sigma_s-\Pi_s)} M_s(\beta)
\label{eq:message:update_special2}
\end{eqnarray}


We will use eq.~\eqref{eq:message:update} (for which the correctness is already proved) for the case when $\alpha\in \Sigma_s$.
We can rewrite it as follows (we use the fact that $\calA_s\cap\Pi_s=\calB_s\cap\Pi_s=\varnothing$):
\begin{equation*}
\hspace{-16pt}M_{s}( \alpha ) := \left[\phi(\alpha)\otimes M_{s-1}(\alpha^-)\right] \oplus
\bigoplus_{\beta\in\widehat\Phi_s(\alpha)\cap \calA_s} M_s(\beta)
\end{equation*}
\begin{equation}
\hspace{30pt}\oplus \bigoplus_{\beta\in\widehat\Phi_s(\alpha)\cap \calB_s} M_s(\beta)
\;\;\oplus\bigoplus_{\beta\in\widehat\Phi_s(\alpha)\cap(\Sigma_s-\Pi_s)} M_s(\beta)
\label{eq:message:update_special3}
\end{equation}
The first and the last terms of the sum in~\eqref{eq:message:update_special3} equal to that
of the sum in~\eqref{eq:message:update_special2}. The lemma below implies
that the same holds for the second and third terms, thus proving the correctness of eq.~\eqref{eq:message:update_special2}.

\begin{lemma} (a) For any $\alpha\in\calA_s$ there holds
\begin{eqnarray}
~\hspace{-95pt} M_s(\alpha)     &=& \phi(\alpha)\otimes W_{s-1}(\alpha^-)
\end{eqnarray}
(b) For any $\alpha\in\calB_s$ there holds
\begin{eqnarray}
~\hspace{-10pt} M_s(\alpha) &=& [\phi(\alpha)\otimes V_{s-1}(\alpha^-,{\alpha_\downarrow}^-)] \oplus \overline{M}_{s}\left({\alpha_\downarrow}\right) \quad
\label{eq:nonspec2}
\end{eqnarray}
\end{lemma}
\begin{proof}
\myparagraph{Part (a)} Suppose that $\alpha\in\calA_s$.
This means that
there are no patterns $\beta\in\Pi^\circ_s$ of the form $\beta=+\alpha$
(recall that $\Pi^\circ_s\subseteq\Sigma^\circ_s$). This in turn implies
that $M_s(\alpha)=W_s(\alpha)$.
This also implies that for any $x\in\calX_s(\alpha)$ there holds $f(x)=f(x^-)$,
and consequently $W_s(\alpha)=\phi(\alpha)\otimes W_{s-1}(\alpha^-)$.

\myparagraph{Part (b)} Suppose that $\alpha\in\calB_s$.
%
%
The definition of $\calB_s$ implies that set $\widehat T_{s}(\alpha)-\widehat T_{s}(\alpha_\downarrow)$ does not contain nodes in $\Pi_s$
or in $\Pi^\circ_s$.
Using this fact and the definition of sets $\calX_s(\cdot)$, $\calX_s(\cdot;\cdot)$ we get the following.
\begin{list}{$\bullet$}{\leftmargin=1em \itemindent=0em \itemsep=0pt}
\item[(i)] If $\alpha_\downarrow\in\Pi_s$ then $\calX_s(\alpha;\Pi_s)=\calX_s(\alpha)-\calX_s(\alpha_\downarrow)$.
\item[(ii)] If $\alpha_\downarrow\notin\Pi_s$ then $\calX_s(\alpha;\Pi_s)$ is a disjoint union of $\calX_s(\alpha)-\calX_s(\alpha_\downarrow)$
and $\calX_s(\alpha_\downarrow;\Pi_s)$.
\item[(iii)] Partial labeling $x\in\calX_s(\alpha)-\calX_s(\alpha_\downarrow)$
cannot end with a pattern $\beta=+\alpha\in\Pi^\circ_s$.
(If such $\beta$ exists then from the fact above we get $\beta\in\widehat T_s(\alpha_\downarrow)$,
i.e.\ $\beta=\ast\alpha_\downarrow$ and $x\in\calX_s(\alpha_\downarrow)$ - a contradiction.)
Therefore, for such $x$ we have $f(x)=\phi(\alpha)\otimes f(x^-)$.
This implies that
 $\sum_{x\in\calX_s(\alpha)-\calX_s(\alpha_\downarrow)}f(x)=\phi(\alpha)\otimes V_{s-1}(\alpha^-,{\alpha_\downarrow}^-)$.
\end{list}

Recall that $M_{s}( \alpha )=\sum_{x\in\calX_s(\alpha;\Pi_s)}f(x)$.
Using this fact and properties (i)-(iii), we conclude that~\eqref{eq:nonspec2}
holds in each of the two cases ($\alpha_\downarrow\!\in\!\Pi_s$ and $\alpha_\downarrow\!\notin\!\Pi_s$).
\end{proof}

\subsection{Proof of Theorem~\ref{th:LlogL}(b)}
For a node $\alpha\in\widehat\Pi_s$ we denote $T^\circ_s(\alpha)=\widehat T_s(\alpha)\cap\Sigma^\circ_s$.

Let us consider the process of a breadth-first search in the tree $G[\widehat\Pi_s]$ starting from the root.
At each step we will keep a certain set of nodes of the tree (which we call active nodes), and the transition to the next step is made by choosing one of the active nodes and replacing it with its children. The process stops when the set of active nodes becomes equal to the set of the leaves of the tree.
To each step of the process we correspond a partition of the set $\Sigma^\circ_s$. The partition is defined by the following rule: if $\alpha_1, \ldots, \alpha_k$ are active nodes, then the partition is
$\Sigma^\circ_s = \bigcup\limits_{\ell=1}^{k} T^\circ_s(\alpha_\ell)\bigcup\limits_{\alpha\in \Sigma^\circ_s, \alpha\notin T^\circ_s(\alpha_\ell),\ell=\overline{1,k}} \left\{\alpha\right\}$.
Let us denote the partition at step $t$ as $D_t$.

Partitions of the set $\Pi_s$ is a poset with respect to the natural order defined as follows: $\left\{S_i\right\}_{i\in I}\leq \left\{P_j\right\}_{j\in J}$ if for any $i\in I$ there is $j\in J$ such that
$S_i\subseteq P_j$.
It is easy to see that $D_0 \geq D_1 \geq D_2 \geq \ldots$. Moreover, if a chosen active node $\alpha$ at step $t$ is a special one and does not belong to $\Sigma^\circ_s$ then $D_t > D_{t+1}$. Indeed, there are at least two children of $\alpha$, denoted as $\alpha_1$ and $\alpha_2$,
such that sets $T^\circ_s(\alpha_1)$ and $T^\circ_s(\alpha_2)$ are nonempty (by the definition of a special node).
As step $t+1$ these sets are separate components of partition $D_{t+1}$,
but at step $t$ these sets still belong to the same component of $D_t$; thus, $D_t\ne D_{t+1}$.

We know that the length of a chain $D_{t_1} > D_{t_2} > \ldots$ cannot exceed $|\Sigma^\circ_s|$.
We conclude that the number of special patterns that do not belong to $\Sigma^\circ_s$ is bounded by $|\Sigma^\circ_s|-1$; this implies Theorem~\ref{th:LlogL}(b).

\subsection{Proof of Theorem~\ref{th:LlogL}(c)}
For brevity denote $t=s-1$, and define a set of pairs
$$
J=\{(\alpha,\beta)\:|\:\alpha,\beta\in\Pi_t,\beta\mbox{ is a strict descendant of $\alpha$ in }G[\Pi_t]\}
$$
Note that $E[\Pi_t]\subseteq J$. In this section we describe a $O(\Pproper\log h)$ preprocessing which will
allow computing values $V_t(\alpha,\beta)$ for any $(\alpha,\beta)\in J$ in $O(\log h)$ time.
The procedure will be based on the following observation;
it follows trivially from the definition~\eqref{eq:LlogL:V}.
\begin{lemma}
For any $(\alpha,\beta)\in J$ there holds
\begin{equation}
V_t(\alpha,\beta)=\bigoplus_{(\alpha',\beta')\in\calP_t(\alpha,\beta)} V_t(\alpha',\beta')
\end{equation}
\end{lemma}
During preprocessing we will compute values $V_t(\alpha,\beta)$ for pairs $(\alpha,\beta)$ in
a certain set $\widetilde J\subseteq J$ of size $O(\Pproper\log h)$. In order to define $\widetilde J$, we need some notation.
For a pattern $\alpha\in\Pi_t$ let $h_\alpha=|\calP_t(\varepsilon_t,\alpha)|$ be the height of $\alpha$ in $G[\Pi_t]$,
and for an integer $d\in[0,h_\alpha]$ let $\alpha^{\uparrow d}$ be the node
of tree $G[\Pi_t]$ obtained from $\alpha$ by taking $d$ steps towards the root. We now define
\begin{eqnarray*}
\widetilde J=\{(\alpha^{\uparrow d},\alpha)\:|\:\alpha\in\Pi_t,d\in[0,h_\alpha],d=2^r\mbox{ for }r\in\mathbb Z_{\ge 0}\}
\end{eqnarray*}

The preprocessing will consist of 3 steps.

\noindent{\bf Step 1}: compute values $W_t(\alpha)$ for all $\alpha\in\Pi_t$.
We do it by traversing nodes $\alpha\in\Pi_t$ of tree $G[\Pi_t]$
and setting
\begin{equation}
W_t(\alpha):=M_t(\alpha)\oplus\bigoplus_{(\alpha,\beta)\in E[\Pi_t]} W_t(\beta)
\end{equation}
This takes $O(\Pproper)$ time.

\noindent{\bf Step 2}: go through $\alpha\in\Pi_t$ and compute
values $V_t(\alpha,\beta)$ for all $\beta\in \Phi_t(\alpha)$
where $\Phi_t(\alpha)$ is the set of children of $\alpha$ in $G[\Pi_t]$.

A naive way is to use the formula
$$
V_t(\alpha,\beta)=M_t(\alpha)\oplus\bigoplus_{\gamma\in\Phi_t(\alpha)-\{\beta\}} W_t(\gamma)
$$
for all $\beta\in\Gamma_t(\alpha)$; however, this would take $O(k^2)$ time where $k=|\Phi_t(\alpha)|$.
Instead, we do the following. Let us order patterns in $\Phi_t(\alpha)$ arbitrarily:
$\Phi_t(\alpha)=\{\beta_1,\ldots,\beta_k\}$. For $i\in[1,k]$ denote
$$
\overrightarrow S_i=\bigoplus_{j=1}^{i-1} W_t(\beta_j) \qquad
\overleftarrow S_i=\bigoplus_{j=i+1}^{k} W_t(\beta_j)
$$
We compute these values in $O(k)$ time by setting $\overrightarrow S_1:=\mathbb{O}$,
$\overleftarrow S_{k}:=\mathbb{O}$
and then using recursions
$$
\overrightarrow S_{i+1}:=\overrightarrow S_{i}\oplus W_t(\beta_{i})\qquad
\overleftarrow S_{i-1}:=\overleftarrow S_{i}\oplus W_t(\beta_{i})
$$
After that we set
$$
V_t(\alpha,\beta_i):=M_t(\alpha)\oplus\overrightarrow S_i\oplus\overleftarrow S_i
$$
For a given $\alpha\in\Pi_t$ the procedure takes $O(|\Phi_t(\alpha)|)$ time,
and thus for all $\alpha\in\Pi_t$ it takes $O(\Pproper)$ time.

We now have values $V_t(\alpha,\beta)$ for all $(\alpha,\beta)\in E[\Pi_t]$.

\noindent{\bf Step 3}: compute values $V_t(\alpha,\beta)$ for all $(\alpha,\beta)\in \widetilde J$
using the recursion
$$
V_t(\alpha^{\uparrow 2d},\alpha):=V_t(\alpha^{\uparrow 2d},\alpha^{\uparrow d})+V_t(\alpha^{\uparrow d},\alpha)
$$
for $d=2^0,2^1,\ldots,2^r,\ldots$ and $(\alpha^{\uparrow 2d},\alpha)\in \widetilde J$.

\myparagraph{Evaluating queries for $(\alpha,\beta)\in J$} We showed how to compute values $V_t(\alpha,\beta)$ for $(\alpha,\beta)\in \widetilde J$ in time $O(\Pproper\log h)$;
let us now describe how to compute
value $V_t(\alpha,\beta)$ for a given $(\alpha,\beta)\in J$ in time $O(\log h)$.
Let us construct a sequence $\beta_0,\beta_1,\ldots,$ as follows:
$\beta_0=\beta$, and for $i\ge 0$ let $\beta_{i+1}=\beta_i^{\uparrow d}$ where
$d$ is the maximum value such that $d=2^r$ for $r\in\mathbb Z_{\ge 0}$ and $\beta_i^{\uparrow d}$ is still
a descendant of $\alpha$. We stop when we get $\beta_k=\alpha$; clearly, this happens after $k=O(\log h)$ steps.
We now set
$$
V_t(\alpha,\beta):=\bigoplus_{i=0}^{k-1} V_t(\beta_{i+1},\beta_i)
$$

\subsection{Proof of Theorem~\ref{th:LlogL}(d)}

We will consider the general case of a commutative semiring and the case when $(R,\oplus,\otimes)\!=\!(\overline{\mathbb R},\min,+)$;
the latter will be called the {\em MAP case}.

Let $d_\alpha$ for $\alpha\in\Pi_{s-1}$ be the number of children of
$\alpha$
in $G[\Pi_{s-1}]$, $d_{\max}=\max_{\alpha\in\Pi_{s-1}}d_\alpha$,
and $\widetilde d_\alpha$ for $\alpha\in\Sigma_s$ be the number
of children of $\alpha$ in $G[\Sigma_s]$. We will present a
$O(\sum_{\alpha\in\Pi_{s-1}}d_\alpha \log d_\alpha)$ preprocessing
technique that will allow computing value $A_s(\alpha)$ for
$\alpha\in\Sigma_s-\{\varepsilon_s\}$
in time $O((\widetilde d_\alpha+1) \log d_{\alpha^-})$. The resulting
complexity will be
$$
\mbox{$
O(\sum\limits_{\alpha\in\Pi_{s-1}}d_\alpha
\log d_{\max})
+\sum\limits_{\alpha\in\Sigma_s}
O((\widetilde d_\alpha+1) \log d_{\max})=O(\Pall\log d_{\max})
$
}
$$
In the {\em MAP case} (i.e.\ when $(R,\oplus,\otimes)\!=\!(\overline{\mathbb R},\min,+)$)
we will present a faster solution. Namely, the preprocessing will take $O(\sum_{\alpha\in\Pi_{s-1}}d_\alpha)=O(\Pproper)$ time,
and computing value $A_s(\alpha)$ for
$\alpha\in\Sigma_s-\{\varepsilon_s\}$ will take
$O(\widetilde d_\alpha+1)$ time, leading to the overall complexity $O(\Pall)$.
For that we will use the {\em Range Minimum Query (RMQ)} problem
which is defined as follows: given $N$ numbers $z_1,\ldots,z_N$,
compute $\min_{k\in I}z_k$ for a given interval $I=[i,j]\subseteq[1,N]$.
It is known~\cite{BerkmanVishkin:93} that with an $O(N)$ preprocessing each
query for can be answered in $O(1)$ time per interval.

As in the previous section, we denote $t=s-1$.
We assume that for each $\alpha\in\Pi_t$
we already have values $W_t(\alpha)$ for all $\alpha\in\Pi_t$
(they were computed in the previous section).

\myparagraph{Preprocessing} Consider $\alpha\in\Pi_t$, and let us fix an ordering
of children of $\alpha$: $\Phi_t(\alpha)=\{\beta_1,\ldots,\beta_d\}$
where $d=d_\alpha$.
For an interval $I\subseteq[1,d]$ we denote
\begin{equation}
S_{I}(\alpha)=\bigoplus_{i\in I} W_t(\beta_i)
\end{equation}
The goal of preprocessing is to build a data structure that
we will allow an efficient computation of $S_I(\alpha)$ for any given interval $I$.

In the MAP case we simply run the preprocessing of~\cite{BerkmanVishkin:93} for
the sequence $W_t(\beta_1),\ldots,W_t(\beta_d)$; this takes $O(d)$ time.
Value $S_I(\alpha)$ for an interval $I$ can then be computed in $O(1)$ time.

In the general case we do the following.
Define a set of intervals
$$
J_d=\{[i,j]\subseteq[1,d]\:|\:j-i=2^r-1,r\in\mathbb Z_{\ge 0}\}
$$
Note that $|J_d|=O(d\log d)$.
We compute quantities
$
S_{I}(\alpha)
$
for all $I\in J_d$. This can be done in $O(d\log d)$ time by
setting $S_{[i,i]}(\alpha):=W_t(\beta_i)$ for $i\in[1,d]$ and then using
recursions
$$
S_{[i,i+2\delta-1]}(\alpha)=S_{[i,i+\delta-1]}(\alpha)\oplus
S_{[i+\delta,i+2\delta-1]}(\alpha)
$$
for $\delta=2^0,2^1,\ldots$.

Value $S_I(\alpha)$ for an interval $I$ can now be
computed in $O(\log d)$ time. Indeed, we can represent $I$
as a disjoint union of $m=O(\log d)$ intervals from $J_d$:
$I=\bigcup_{i=1}^{m}I_i$ with $I_i\in J_d$. We can then use the formula
\begin{equation}
S_{I}(\alpha)=\bigoplus_{i=1}^{m}S_{I_i}(\alpha)
\end{equation}

\myparagraph{Computing $A_s(\alpha)$ for $\alpha\in\Sigma_s-\{\varepsilon_s\}$}
Denote $d=d_{\alpha^-}$.
As discussed earlier, $\widehat \Phi_s(\alpha)$ (the set of children
of $\alpha$ in $G[\widehat\Pi_s]$) is isomorphic to
$\Phi_{s-1}(\alpha^-)$.
Let $\widehat \Gamma_s(\alpha)=\{\beta_1,\ldots,\beta_d\}$ be the ordering
of patterns in $\widehat \Phi_s(\alpha)$ such that
$\beta_1^-,\ldots,\beta_d^-$ is the ordering of patterns in
$\Phi_{s-1}(\alpha^-)$
chosen in the preprocessing step. We need to compute
$$
A_s(\alpha)=\bigoplus_{i\in J}W_{s-1}(\beta_i^-)\;,\quad
J\triangleq\{i\in[1,d]\:|\:\beta_i\in\calA_s\}
$$
Denote $\overline J=[1,d]-J$.
We can represent $J$ as a disjoint union of at most $|\overline J|+1$ intervals
$I_1,\ldots,I_m$
where $I_i\subseteq[1,p]$.
Clearly, $|\overline J|=\widetilde d_\alpha$ (see Fig.~\ref{fig:Sigmas}),
and so $m\le \widetilde d_\alpha+1$.
We can write
\begin{equation}
A_s(\alpha)=\bigoplus_{i=1}^{m} S_{I_i}(\alpha^-)
\label{eq:LlogL:computingAs:a}
\end{equation}
As discussed above, each value $S_{I_i}(\alpha^-)$ can be computed
in $O(1)$ time in the MAP case and in $O(\log d)$ in the general case.
Thus, computing $A_s(\alpha)$ takes respectively $O(\widetilde d_\alpha+1)$
and $O((\widetilde d_{\alpha}+1)\log d_{\alpha^-})$ time, as desired.

%

\else
\fi

\section{MAP for non-positive costs}\label{sec:negative}
In this section we assume that $(R,\oplus,\otimes)\!=\!(\overline{\mathbb R},$ $\min,+)$
and $c_\alpha\le 0$ for all $\alpha\in\Pi^\circ$.
\cite{Komodakis:CVPR09} gave an algorithm that makes $\Theta(n\Pall)$ comparisons
and $\Theta(n\Pall)$ additions. We will present a modification that
makes only $O(n|I(\Gamma)|)$ comparisons. The number of additions
in general will still be $O(n\Pall)$, but we will show that
in certain scenarios it can be reduced using a Fast Fourier Transform (FFT).

We will assume that $\Gamma$ contains at least one word $\alpha$ with $|\alpha|=1$
(it can always be added if needed).

As usual, we first select a set of patterns $\Pi$ with $\Pi_0=\{\varepsilon_0\}$;
this step will be described later.
For a pattern $\alpha\in \Pi$ let $\alpha^\leftarrow$ be the longest proper prefix of $\alpha$ that is in $\Pi$
($\alpha=\alpha^\leftarrow+$, $\alpha^\leftarrow\in\Pi$). If $\Pi$ does not contains proper prefixes of $\alpha$
then $\alpha^\leftarrow$ is undefined.

\ifTR
We can now present the algorithm.
\fi
\begin{algorithm}
\caption{Computing $Z=\min\limits_{x\in D^{1:n}} f(x)$ (if $c_\alpha\le 0$)}\label{alg:DP}
\begin{algorithmic}[1]
\STATE initialize messages: set $M_0(\varepsilon_0):=0$
\STATE for $s=1,\ldots,n$ traverse nodes $\alpha\in \Pi_s$ of forest $G[\Pi_s]$
starting from the leaves and set
\begin{equation}
M_{s}( \alpha ) := \min\{M_{p}( \alpha^\leftarrow  )
+ \psi(\alpha),
\min\limits_{(\alpha,\beta)\in E[\Pi_s]} M_{s}( \beta)\}
\label{eq:message:update3}
\end{equation}
where $p$ is the end position of $\alpha^\leftarrow$ ($\alpha^\leftarrow=([\cdot,p],\cdot))$
and $\psi(\alpha)=f(\alpha)-f(\alpha^\leftarrow)$.
If $\alpha^\leftarrow$ is undefined then ignore the first expression in~\eqref{eq:message:update3}.
\STATE return
$
Z:=\min_{\alpha\in \Pi_n}  M_n(\alpha)
$
\end{algorithmic}
\end{algorithm}

\myparagraph{Selecting $\Pi$} It remains to specify how to choose set $\Pi$. 
For patterns $\alpha,\beta,\gamma$ we define
\begin{equation}
\langle\alpha|\beta|\gamma\rangle
=\{u=+\beta+\:|\:\alpha\beta\gamma=\ast u\ast\}
\label{eq:DP:LRangle}
\end{equation}
%
\begin{theorem} Suppose that $\Pi_0=\{\varepsilon_0\}$ and set $\Pi$ contains set
\begin{eqnarray}
\widetilde\Pi=\left\{\beta
\begin{picture}(1,0)
  \put(3,-10){\line(0,1){24}}
\end{picture}
\begin{tabular}{l}
$\exists\mbox{ labeling }x\alpha\beta\gamma y\in D^{1:n}$ s.t.\ \\
 (a) $\alpha\beta,\beta\gamma\in\Pi^\circ$; (b) $\langle x\alpha|\beta|\gamma y\rangle\cap\Pi^\circ=\varnothing$
\end{tabular}
\hspace{-7pt}\right\}
\label{eq:Idelta}
\end{eqnarray}
Then Alg.~\ref{alg:DP} returns the correct value of $Z\!=\!\!\min\limits_{x\in D^{1:n}}\! f(x)$.\!\!\!\!\!\!
\label{th:correctness}
\end{theorem}

\ifTR
A proof of this theorem is given in Sec.~\ref{sec:algOne:correctness}.
\fi

A simple valid option is to set
$\Pi=\{\alpha\:|\:\exists\alpha\ast,\ast\alpha\in\Pi^\circ\}$.
Computing set $\widetilde\Pi$ is slightly more complicated, but can
still be done in polynomial time for a given $\Pi$ (we omit this procedure).
In order to analyze set $\widetilde\Pi$, let us define
\begin{equation*}
I_\delta=\left\{\beta
\begin{picture}(1,0)
  \put(3,-10){\line(0,1){24}}
\end{picture}
\begin{tabular}{l}
$\exists\mbox{ word }x\alpha\beta\gamma y\mbox{ with $|x|=|y|=\delta$}$ \\
s.t. (a) $\alpha\beta,\beta\gamma\in\Gamma$; (b) $\langle x\alpha|\beta|\gamma y\rangle\cap\Gamma=\varnothing$ \\
\end{tabular}
\hspace{-7pt}\right\}
\end{equation*}
where set $\langle\cdot|\!\cdot\!|\cdot\rangle$ for words is defined similarly to~\eqref{eq:DP:LRangle}:
\begin{equation}
\langle\alpha|\beta|\gamma\rangle
=\{\hat\alpha\beta\hat\gamma\:|\:\alpha=\ast\hat\alpha,\gamma=\hat\gamma\!\ast\mbox{ and }\hat\alpha,\hat\gamma\ne\varepsilon\}
\end{equation}

As $\delta$ increases, set $I_\delta$ monotonically shrinks, and stops changing after $\delta\ge \ell_{\max}=\max_{\alpha\in\Gamma}|\alpha|$.
We denote this limit set as $I_\infty$, so that $I_\infty\subseteq I_0\subseteq I(\Gamma)\cup\{\varepsilon\}$.
It can be seen that $\widetilde\Pi_s=\{([\cdot,s],\alpha)\:|\:\alpha\in I_\infty\}$ for all $s\in[\ell_{\max},n\!-\!\ell_{\max}\!+\!1]$.

\myparagraph{Complexity} Assume that we use $\Pi=\widetilde\Pi$.
The algorithm performs two types of operations: comparisons
(to compute minima) and arithmetic operations (to compute the first expression in~\eqref{eq:message:update3}).
The number of comparisons does not exceed the total number of edges in graphs $G[\Pi_s]=(\Pi_s,E[\Pi_s])$
for $s\in[1,n]$, which is smaller than the number of nodes (since graphs are forests).
Thus, comparisons take $O(n|I_\infty|)$ time.

The time for arithmetic operations depends on how we compute quantities $f(\alpha)$.
One possible approach is to use Lemma~\ref{lemma:varphis}
for computing $f(\alpha)$ for all $\alpha\in \widehat\Pi$ where
$\widehat\Pi$ is the set of prefixes of patterns in $\Pi^\circ$ (note that $\Pi\subseteq\widehat\Pi$).
We have $|\widehat\Pi_s|\le \Pall+1$, and therefore the resulting overall complexity
is $O(n\Pall)$.
Next, we describe an alternative approach based on a Fast Fourier Transform.


\subsection{Computing $f(\alpha)$ using FFT}
For a word $\alpha$ and index $s\ge |\alpha|$ let $\alpha_s$ be the pattern $([s-|\alpha|+1,s],\alpha)$.
It is easy to see that
\begin{equation*}
f(\alpha_s) = \sum\limits_{\beta\in\Gamma} f_s(\alpha|\beta)\mbox{~~where~~}f_s(\alpha|\beta) =\sum_{t:\alpha_s=\ast\beta_t\ast} c_{\beta_t}
\end{equation*}
\begin{lemma}
For fixed words $\alpha,\beta$ quantities $f_s(\alpha|\beta)$ for $s\in[1,n]$
can be computed in $O(n\log n)$ time.
\end{lemma}
\begin{proof}
We assume that $|\alpha|\ge|\beta|$, otherwise the claim is trivial.
Let $p=|\alpha|-|\beta|+1$, and define sequences $a\in\mathbb R^{n-|\alpha|+1},b\in\mathbb R^{n-|\beta|+1},\lambda\in\{0,1\}^p$ via
\begin{eqnarray*}
a_i&=&f_{i+|\alpha|}(\alpha|\beta) \hspace{50pt} \forall i\in[1,n\!-\!|\alpha|\!\,] \\
b_j&=&c_{([j,j+|\beta|-1],\beta)} \hspace{39.5pt} \forall j\in[1,n\!-\!|\beta|\!+\!1] \\
\lambda_k&=&[\alpha_{k:k+|\beta|-1} = {\beta}] \hspace{29pt} \forall k\in[1,p]
\end{eqnarray*}
where $[\cdot]$ is the Iverson bracket. It can be checked that
$$
a_i=\sum_{k=1}^p b_{i+k} \lambda_k\qquad\forall i\in[1,n\!-\!|\alpha|\,]
$$
Thus, $a$ is the convolution of $b$ and the reverse of $\lambda$.
The convolution of such sequences can be computed in $O(n\log n)$ using a Fast Fourier Transform.
\end{proof}

In practice this method can be useful for computing $f_{s}(\alpha|\beta)$ when $|\alpha|\gg|\beta|$. A natural way to do this is first
choose a subset $S\subseteq \Gamma$ of words that are very often included as subwords in words from $I_{\infty}$ (usually, this
means $S = \left\{\beta\:|\: \beta\in\Gamma, |\beta|\leq \delta\right\}$ where $\delta$ is some threshold constant). Then computing $f_s(\alpha|\beta)$ for all $\alpha\in I_{\infty}$ and
$\beta\in S$ will take $O\left(\left|I_{\infty}\right|\cdot\left|S\right| n\log n\right)$ time.
For $\beta\in\Gamma-S$ quantities $f_{s}(\alpha|\beta)$ can be computed directly.

An example of using such an approach is the following theorem; it is proved by taking $\delta = 1$ and $S = D$.
\begin{theorem} Suppose $\Gamma = D\cup A$ where $A$ consists of words of a fixed length $\ell$. Then Algorithm~\ref{alg:DP} can be implemented  in
$O\left(\left|I_{\infty}\right|\cdot \left|D\right| n \log n\right)$ time.
\end{theorem}


\ifTR
\subsection{Proof of Theorem \ref{th:correctness} (correctness)}\label{sec:algOne:correctness}

First,  let us prove that for all patterns $\alpha=([\cdot,s],\cdot)\in\Pi$
there holds
\begin{equation}
\label{ineq:1}
M_s(\alpha)\ge \min_{x=\ast\alpha\in D^{1:s}} f(x) 
\end{equation}
We use induction on the order used in the algorithm.
The base case $\alpha=\varepsilon_0$ is ensured by the initialization step.
Consider pattern $\alpha\in\Pi-\{\varepsilon_0\}$.
Suppose that $\alpha^\leftarrow$ is defined; let $p$  be the end position of $\alpha^\leftarrow$.
Consider partial labeling $y=\ast\alpha^\leftarrow\in D^{1:p}$, and let $x$ be its unique extension by $s-p$ letters
($x=y+$) such that $x=\ast\alpha$.
Due to non-positivity of costs $c_\beta$ we have
$$
f(x)=f(y) + \psi(\alpha)+
\!\!\!\!\sum_{\substack{\beta=([i,j],\cdot)\in\Pi^\circ \\ x=\ast\beta\ast \\ i\le s-|\alpha|,\;j>p}}\!\!\!\! c_\beta\le f(y) + \psi(\alpha)
$$
Applying the induction hypothesis and the inequality above yields the desired claim:
\begin{eqnarray*}
M_{s}( \alpha ) &=& \min\{M_{p}( \alpha^\leftarrow ) + \psi(\alpha),
\min\limits_{(\alpha,\beta)\in E[\Pi_s]} M_{s}(\beta)\} \\
 &\ge&
\min\{\min_{y=\ast\alpha^\leftarrow}f( y ) + \psi(\alpha),
\min_{(\alpha,\beta)\in E[\Pi_s]}\min_{x=\ast\beta} f(x)\} \\
&\ge&
\min\{\min_{x=\ast\alpha}f( x ) ,
\min_{x=\ast\alpha}f( x )\}
=\min_{x=\ast\alpha}f( x )
\end{eqnarray*}
where in the equations above $x$ always denotes a partial labeling in $D^{1:s}$
and $y$ denotes a partial labeling in $D^{1:p}$. If $\alpha^\leftarrow$ is undefined
then we can write similar inequalities but omitting expressions contaiting $\alpha^\leftarrow$.

By applying the claim to patterns $\alpha=([\cdot,n],\cdot)\in\Pi$ we conclude that $Z\ge \min_{x\in D^{1:n}} f(x)$.
The remainder of this section is devoted to the proof of the reverse inequality: $Z\le \min_{x\in D^{1:n}} f(x)$.

Let us fix $x^\ast\in\arg\min_{x\in D^{1:n}} f(x)$. Let
\begin{equation*}
\Lambda \;=\;\{\alpha\in\Pi^\circ\:|\: x^\ast=\ast\alpha\ast\}
\end{equation*}
be the set of patterns present in $x^\ast$, and
\begin{equation*}
\widehat\Lambda=\{\alpha\in\Lambda\:|\:
\mbox{there is no }\widehat \alpha\in\Lambda-\{\alpha\}\mbox{ \!with\! }\widehat\alpha=\ast\alpha\ast\}
\end{equation*}
be the set of {\em maximal} patterns in $\Lambda$.
We can assume w.l.o.g.\ that for each $k\in[1,n]$ there exists $\alpha=([i,j],\cdot)\in\widehat\Lambda$ with $k\in[i,j]$.
Indeed, if it is not the case for some $k$ then we can modify $x^\ast$ by replacing the $k$-th letter of $x^\ast$
with some letter $c\in\Gamma\cap D$; this operation does not increase $f(x^\ast)$.

We define a total order $\preceq$ on patterns $\alpha=([i,j],\cdot)\in\widehat\Lambda$ as
the lexicographical order with components  $(i,j)$
(the first component is more significant).
\begin{lemma} (a) For each pattern $\beta\in\widehat\Lambda$ there holds $\beta\in \Pi$. \\
(b) Consider two consecutive patterns $\alpha_1\prec \alpha_2$ in $\widehat\Lambda$
with $\alpha_1=([i_1,j_1],\cdot)$, $\alpha_2=([i_2,j_2],\cdot)$, and let
$\beta=x^\ast_{i_2:j_1}$ be the
pattern at which they intersect. (By the assumption above, $i_2 \le j_1 + 1$). There holds $\beta\in \Pi$. \\
(c) For the patterns in (b), condition
 $M_{j_1}(\alpha_1)\le f(x^\ast_{1:j_1})$ implies $M_{j_2}(\alpha_2)\le f(x^\ast_{1:j_2})$.
\label{lemma:correctness}
\end{lemma}
\begin{proof}
\myparagraph{Part (a)} We can write labeling $x^\ast$ as $x^\ast=x\alpha\beta\gamma y$
where patterns $\alpha$, $\beta$ are empty.
Let us show that this choice satisfies conditions in~\eqref{eq:Idelta}. Condition (a) holds
since $\alpha\beta=\beta\gamma=\beta\in\Pi^\circ$. Suppose that (b) does not
hold, then there exists pattern $u=x^\ast_{k:\ell}\in\Pi^\circ$ with $u=+\beta+$.
We have
$u\in\Lambda$
and thus $\beta\notin\widehat\Lambda$ - a contradiction. Therefore, $\beta\in \Pi$.

\myparagraph{Part (b)}
The definitions of $\preceq$ and $\widehat\Lambda$ imply
that $i_1<i_2$. (If $i_1=i_2$ then we must have $j_1<j_2$, but then $\alpha_1\notin\widehat\Lambda$ - a contradiction.)
There also holds $j_1<j_2$ (otherwise we would have  $\alpha_2\notin\widehat\Lambda$ - a contradiction).
This means that we can write labeling $x^\ast$ as $x^\ast=x\alpha\beta\gamma y$ where
$\alpha\beta=\alpha_1$, $\beta\gamma=\alpha_2$ and the pattern intervals are as follows:
\begin{eqnarray*}
&x\hspace{42pt}\alpha\hspace{37pt}\beta\hspace{37pt}\gamma\hspace{42pt}y
& \\
&
 [1,i_1\!-\!1]\quad [i_1,i_2\!-\!1] \quad [i_2,j_1] \quad
 [j_1\!+\!1,j_2] \quad [j_2\!+\!1,n]
&
\end{eqnarray*}
Let us show that this choice satisfies conditions in~\eqref{eq:Idelta}. Condition (a) holds
since $\alpha\beta=\alpha_1\in\Pi^\circ$ and $\beta\gamma=\alpha_2\in\Pi^\circ$. Suppose that (b) does not
hold, then there exists pattern $u=x^\ast_{k:\ell}\in\Pi^\circ$ where
$k<i_2$ and $\ell>j_1$.
This means that $u\in\Lambda$.

From the definition of $\widehat\Lambda$, there exists pattern $\widehat u=x^\ast_{\widehat k:\widehat \ell}\in\widehat\Lambda$
with $[k,\ell]\subseteq[\widehat k,\widehat \ell]$. To summarize, we have $\widehat k\le k<i_2$ and $j_1<\ell\le \widehat \ell$.

If $\widehat k\le i_1$ then $[i_1,j_1]\subset[\widehat k,\widehat \ell]$ and thus $\alpha_1\notin\widehat\Lambda$ - a contradiction.
Thus,
there must hold $\widehat k>i_1$.
Similarly, we prove that  $\widehat\ell<j_2$.
This implies that $\alpha_1\prec \widehat u\prec \alpha_2$,
and therefore patterns $\alpha_1$ and $\alpha_2$ are not consecutive in $\widehat\Lambda$ - a contradiction.

\myparagraph{Part (c)} $\beta$ is a proper prefix of $\alpha$ that belongs to $\Pi$.
Therefore, $\alpha^\leftarrow$ is defined.
%
We can define a sequence of patterns
$\beta_0=\beta,\beta_1,\ldots,\beta_m=\alpha_2$ with $\beta_k=([i_2,s_k],\cdot)$, $s_0=j_1<s_1<\ldots<s_m=j_2$
respectively such that
$\beta_k\in\Pi$ and $\beta_{k-1}=\beta_k^\leftarrow$ for $k\in[1,m]$. Let us prove by
induction on $k$ that $M_{s_k}(\beta_k)\le f(x^\ast_{1:s_k})$ for $k\in[0,m]$.

Let us first check the base of the induction. 
Since $\beta,\alpha_1\in \Pi_{j_1}$ and $\alpha_1=\ast\beta$,
by the definition of graph $G[\Pi_{j_1}]$ there is a (unique) path
$\gamma_0,\gamma_1,\ldots,\gamma_r$ from $\gamma_0\!=\!\beta$ to $\gamma_r\!=\!\alpha_1$
with $(\gamma_l,\gamma_{l+1})\in E[\Pi_{j_1}]$.
%
%
By eq.~\eqref{eq:message:update3},
$M_{j_1}(\gamma_l) \leq M_{j_1}\left( \gamma_{l+1}\right)$. Therefore, $M_{j_1}(\beta) \leq M_{j_1}\left( \alpha_1\right) \leq f(x^\ast_{1:j_1})$
where the last inequality holds by the assumption of part (c). This establishes the base case.

Now suppose that the claim holds for $k-1\in[0,m-1]$; let us prove it for $k$.
Denote $p=s_{k-1}$ and $s=s_k$. Note, $p$ is the index in step 2 of the algorithm
during the processing of $(s,\beta_k)$. From eq.~\eqref{eq:message:update3} and the induction hypothesis
we get
$$
M_s(\beta_k)\le M_p(\beta_{k-1})+\psi(\beta_k)\le F_p(x^\ast_{1:p})+\psi(\beta_k)
$$
We will prove next that $f(x^\ast_{1:p})+\psi(\beta_k)=f(x^\ast_{1:s})$,
thus completing the induction step. It suffices to show that there is no
pattern $\gamma=x^\ast_{i:j}$ with $i<i_2$ and $p<j\le s$.

Suppose on the contrary that such pattern exists. By the definition of $\widehat\Lambda$ there
exists pattern $\widehat\gamma=([\widehat i,\widehat j],\cdot)\in\widehat\Lambda$
with $[i,j]\subseteq[\widehat i,\widehat j]$.
We have $\widehat i\le i<i_2$ and $\widehat j\ge j>p\ge s_0=j_1$.
These facts imply that $\alpha_1\prec\widehat\gamma\prec\alpha_2$,
contradicting the assumption that $\alpha_1$ and $\alpha_2$ are consecutive
patterns in $\widehat\Lambda$.
\end{proof}

Lemma~\ref{lemma:correctness} implies the main claim.
\begin{corollary}
For each $\alpha=([i,j],\cdot)\in\widehat\Lambda$
there holds $M_j(\alpha)\le f(x^\ast_{1:j})$, and therefore $Z\le f(x^\ast)$.
\end{corollary}
\begin{proof}
We use induction on the total order $\preceq$. The lowest pattern
$\alpha\in\widehat\Lambda$ starts at position 1;
for such pattern the claim follows by inspecting Algorithm \ref{alg:DP}.
The induction step follows from Lemma~\ref{lemma:correctness}(c).
\end{proof}
\else
\fi

\section*{Acknowledgements}
We thank Herbert Edelsbrunner for helpful discussions.


\bibliographystyle{plain}
\bibliography{ICML13}

\end{document}